\documentclass{article} % For LaTeX2e
\usepackage[preprint]{colm2026_conference}

\usepackage[T1]{fontenc}
\usepackage{microtype}
\usepackage{hyperref}
\usepackage{url}
\usepackage{booktabs}
\usepackage{tabularx}
\usepackage{multirow}
\usepackage{amsmath}
\usepackage{amssymb}
\usepackage[most]{tcolorbox}
\usepackage{enumitem}
\usepackage{ragged2e} % For better text wrapping
\usepackage{adjustbox}
\usepackage[table]{xcolor}
\usepackage{siunitx}
\usepackage{xfp}
\usepackage{graphicx}
\usepackage{wrapfig}
\usepackage{float}
\usepackage{algorithm}
\usepackage{algpseudocode}

\definecolor{tacinfo}{RGB}{59,130,246}
\definecolor{tacasst}{RGB}{139,92,246}
\definecolor{tacadv}{RGB}{245,158,11}
\definecolor{tacval}{RGB}{16,185,129}
\definecolor{tacemo}{RGB}{239,68,68}
\definecolor{tacpara}{RGB}{6,182,212}
\definecolor{tacself}{RGB}{236,72,153}
\definecolor{tacques}{RGB}{99,102,241}
\definecolor{tacreap}{RGB}{249,115,22}
\definecolor{tacemp}{RGB}{20,184,166}

\definecolor{tableaccent}{RGB}{67,104,173}
\colorlet{tableoursbg}{tableaccent!4}
\colorlet{tablebestbg}{tableaccent!18}
\definecolor{stickaccent}{RGB}{39,121,107}
\colorlet{stickbestbg}{stickaccent!18}
\definecolor{promptbg}{RGB}{255,255,255}
\definecolor{qualitybg}{RGB}{249,249,249}
\definecolor{tokenbg}{RGB}{244,244,244}
\newcommand{\sym}[1]{\textsuperscript{#1}}
\definecolor{sigloss}{RGB}{95,95,95}
\newcommand{\sigworse}[1]{\textcolor{sigloss}{#1}}
\newcommand{\losssym}[1]{\textsuperscript{\textcolor{sigloss}{#1}}}
\newcommand{\siglosscell}{\color{sigloss}}

\usepackage{lineno}

\definecolor{darkblue}{rgb}{0, 0, 0.5}
\hypersetup{colorlinks=true, citecolor=darkblue, linkcolor=darkblue, urlcolor=darkblue}

\newcommand{\ourframework}{\texorpdfstring{\textsc{Mint}}{MINT}}
\newcommand{\minticon}{\texorpdfstring{\hspace{0.15em}\raisebox{-0.15em}{\includegraphics[height=1em]{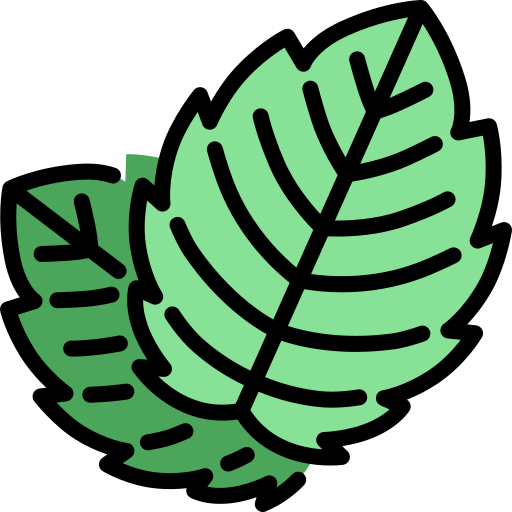}}}{}}
\DeclareFontShape{T1}{ppl}{m}{scit}{<->ssub*ppl/m/sc}{}

\newcommand{\secref}[1]{\S\ref{#1}}
\title{Discourse Diversity in Multi-Turn Empathic Dialogue}

\author{Hongli Zhan$^1$ \quad Emma S. Gueorguieva$^1$ \quad Javier Hernandez$^2$ \quad Jina Suh$^{3}$\thanks{Work done while at Microsoft Research.}\\
    \textbf{Desmond C. Ong$^1$ \quad Junyi Jessy Li$^1$} \\
    $^1$The University of Texas at Austin \\
    $^2$Microsoft Research \quad $^3$University of Washington \\
    \color{blue}\footnotesize\texttt{honglizhan@utexas.edu \quad jessy@utexas.edu}
}

\begin{document}

\ifcolmsubmission
\linenumbers
\fi

\maketitle

\begin{center}
\vspace{-2.3em}
\small
\href{https://honglizhan.github.io/mint-empathy/}{\raisebox{-0.15em}{\includegraphics[height=1.05em]{mint.png}}\,\textbf{Project Page}}
\hspace{0.8em}
\href{https://github.com/honglizhan/mint-empathy}{\raisebox{-0.3em}{\includegraphics[height=1.4em]{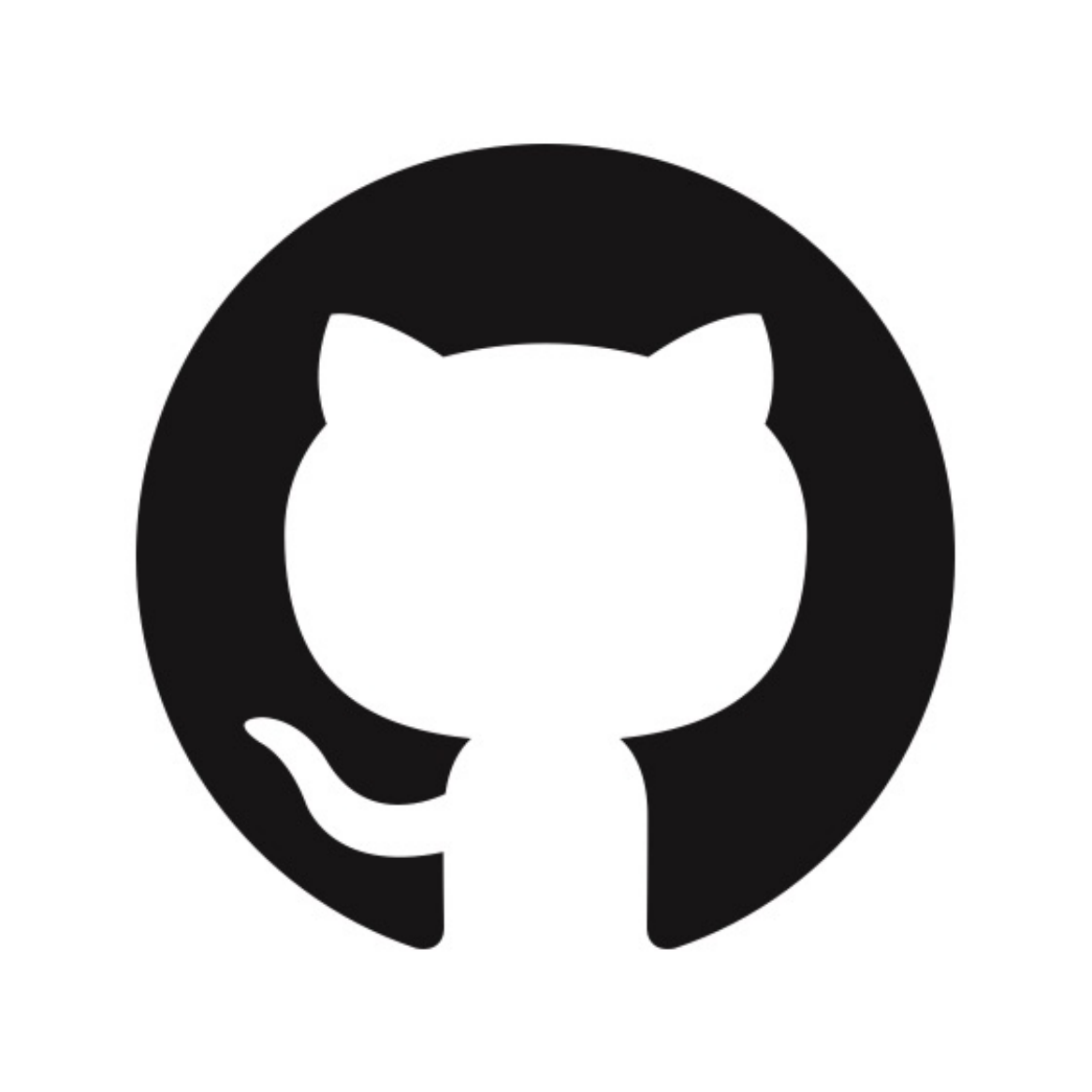}}\,\textbf{Code \& Data}}
\hspace{0.8em}
\href{https://huggingface.co/hongli-zhan/empathy-tactic-taggers-llama3.1-8b}{\raisebox{-0.35em}{\includegraphics[height=1.5em]{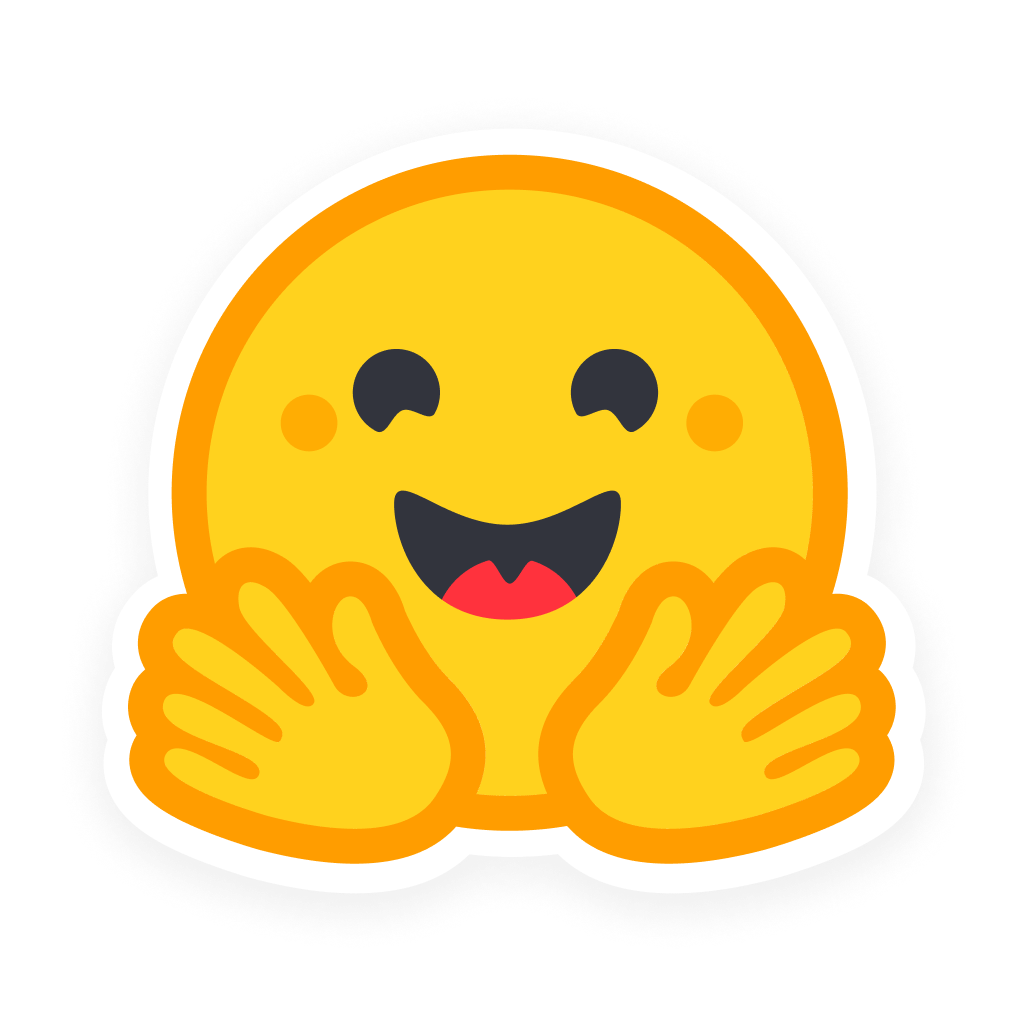}}\,\textbf{Taggers}}
\hspace{0.8em}
\href{https://huggingface.co/hongli-zhan/MINT-empathy-Qwen3-1.7B}{\raisebox{-0.35em}{\includegraphics[height=1.5em]{figures/hf-logo.png}}\,\textbf{Models}}
\vspace{0.3em}
\end{center}

\begin{abstract}
Large language models (LLMs) produce responses rated as highly empathic in single-turn settings \citep{ayers2023comparing, lee2024LLM}, yet they are also known to be formulaic generators that reuse the same lexical patterns, syntactic templates, and discourse structures across tasks \citep{jiang2025artificialhivemind, shaib-etal-2024-detection, namuduri2025qudsim}. Less attention has been paid to whether this formulaicity extends to the level of \emph{discourse moves}, i.e., what a response \emph{does} for the person it is addressing. This question is especially consequential for empathic dialogue, where effective support demands not just a kind response at one moment but varied strategies as a conversation unfolds \citep{stiles1998responsiveness}. Indeed, prior work shows that LLMs already reuse the same tactic sequences more than human supporters in single-turn settings \citep{gueorguieva2026large}. We extend this analysis to multi-turn conversations and find that the rigidity compounds: once a tactic appears in a supporter turn, LLMs reuse it in the next at nearly double the rate of humans ($0.50$--$0.56$ vs.\ $0.27$). This pattern holds across LLMs serving as natural supporters in real emotional support conversations, and is invisible to standard similarity metrics. To address this gap, we introduce \ourframework{}\minticon{} (\textbf{M}ulti-turn \textbf{I}nter-tactic \textbf{N}ovelty \textbf{T}raining), the first reinforcement learning framework to optimize discourse move diversity across multi-turn empathic dialogue. The best \ourframework{} variant combines an empathy quality reward with a cross-turn tactic novelty signal, improving aggregate empathy by an average of $25.3\%$ over vanilla across $1.7$B and $4$B models while reducing cross-turn discourse move repetition by $26.3\%$ on the $4$B model, surpassing all baselines including quality-only and token-level diversity methods on both measures. These results suggest that what current models lack is not empathy itself, but the ability to vary their discourse moves across the arc of a conversation. We publicly release our data and code at \url{https://github.com/honglizhan/mint-empathy}.
\end{abstract}

\section{Introduction}
\label{sec:intro}

\begin{figure*}[t!]
    \centering
    \includegraphics[width=\textwidth]{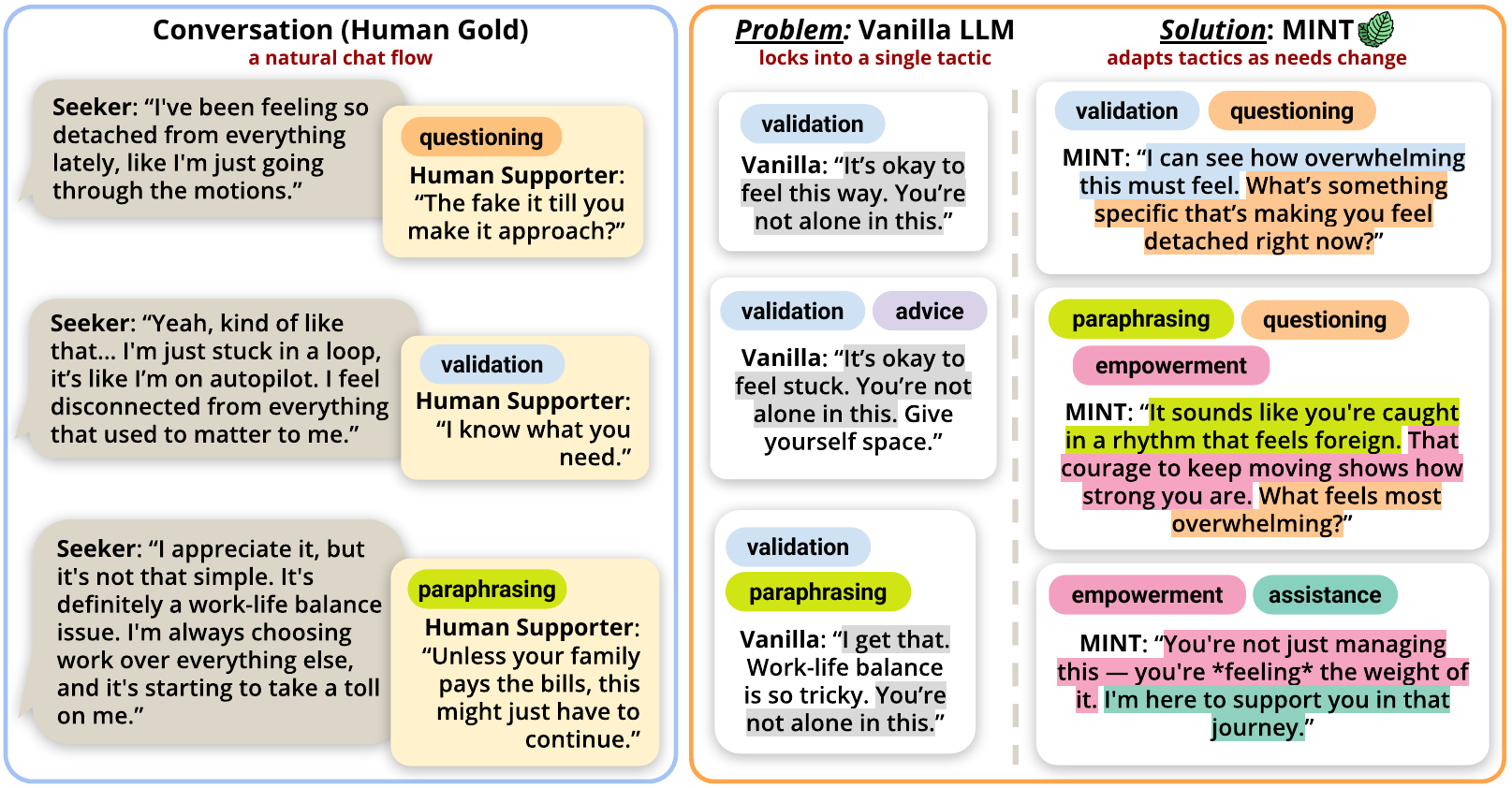}
    \vspace{-2em}
    \caption{As the seeker's needs evolve, vanilla LLMs recycle a narrow tactic set, whereas \ourframework{}\minticon{} adapts its discourse moves across turns.}
    \label{fig:teaser}
    \vspace{-1em}
\end{figure*}

Empathic support plays a central role in how people navigate emotional distress \citep{rogers1957necessary, zaki2014empathy}. LLMs are increasingly capable empathic responders: people consistently rate AI-generated supportive responses as more empathic than human-written ones \citep{lee2024LLM, zhan2024resort, li2025exploring, rubin2025comparing, ong2025ai}, even when the human responses come from trained professionals \citep{ovsyannikova2025third}. But empathy over the course of a conversation requires more than individually warm messages --- it demands that the supporter adapt their approach as new concerns surface \citep{stiles1998responsiveness}. Sustained empathy is less about warmth in any single message than about the variation of approach across the arc of a conversation.

Yet this is precisely where current LLMs fall short: they are increasingly understood to be formulaic generators. At the surface level, they produce homogeneous outputs across models and samples \citep{jiang2025artificialhivemind}, reuse syntactic templates \citep{shaib-etal-2024-detection}, and follow similar discourse structures even when content differs \citep{namuduri2025qudsim}. While these findings document repetition at the lexical, syntactic, and structural levels, less attention has been paid to whether LLMs are also repetitive at the level of \emph{discourse functions} --- what a response \emph{does} for a seeker, such as validating feelings, asking questions, or offering reappraisal. \cite{gueorguieva2026large} took a first step by developing a taxonomy of empathy tactics and showing that LLMs reuse the same tactic sequences at substantially higher rates than human supporters, though their analysis was limited to single-turn settings.

We extend this finding to multi-turn conversations (\secref{subsec:tactic-problem}), where the problem compounds. Using a metric we call \emph{tactic stickiness}, the probability that a tactic used in one turn reappears in the next, we show that LLMs reuse tactics at nearly double the rate of human supporters ($0.50$--$0.56$ vs.\ $0.27$), a pattern consistent across models spanning multiple families. While lexical metrics pick up some surface-level phrase reuse between consecutive turns, semantic similarity fails to detect any difference at all, and neither captures repetition at the level of discourse functions. This has direct practical consequences: without the ability to vary their discourse moves across a conversation, LLMs cannot credibly serve as tools for sustained emotional support, despite promising evidence of reductions in loneliness \citep{de2025ai}, suicidal ideation \citep{maples2024loneliness}, and depression symptoms \citep{heinz2025randomized}.

We address this gap with \ourframework{}\minticon{} (\textbf{M}ulti-turn \textbf{I}nter-tactic \textbf{N}ovelty \textbf{T}raining), a reinforcement learning framework that, to our knowledge, is the first to \textbf{optimize diversity at the level of discourse moves across multi-turn} empathic dialogue. Existing work on empathic dialogue has optimized what the model says at each turn --- whether through quality rewards \citep{zhou-etal-2023-facilitating, wang2025rlverreinforcementlearningverifiable} or strategy selection \citep{liu-etal-2021-towards, tu-etal-2022-misc, cheng-etal-2022-improving} --- but not how the model's discourse moves evolve across turns. The key insight behind \ourframework{} is that tactic diversity should not be defined in a vacuum: a response that uses \textit{validation}, \textit{advice}, and \textit{questioning} is tactically broad on its own, but if the previous turn used the same three tactics, the conversation as a whole still feels repetitive. \ourframework{} therefore compares each turn's tactic profile against the preceding turn and rewards departures from the established pattern, while anchoring diversity to a base empathy quality reward so that tactical variation is only credited when genuinely supportive (\secref{sec:method}). Figure~\ref{fig:teaser} illustrates this contrast: a vanilla LLM locks into the same tactics across turns, while \ourframework{} shifts fluidly as the seeker's needs evolve. This design reflects the central hypothesis of our work: \emph{what current models lack is not empathy itself, but the ability to strategically vary their discourse moves across the arc of a conversation.}

We evaluate on two complementary dimensions: \emph{empathy quality}, via the Lend-an-Ear framework \citep{kumar2025large} adapted into a turn-level protocol that achieves expert-level agreement ($\kappa_w = 0.58$; \secref{subsec:eval}); and \emph{tactic diversity}, via the stickiness metric introduced in \secref{subsec:tactic-problem}. On $315$ held-out supporter turns across $50$ conversations, \ourframework{} improves aggregate empathy by $25.3\%$ over Vanilla prompting across both $1.7$B and $4$B models, while reducing tactic stickiness by $26.3\%$ on the $4$B model, surpassing all baselines including quality-only RL and token-level diversity methods. In sum, this work offers three contributions: (1)~the first multi-turn evidence that discourse move repetition, previously documented only in single-turn settings \citep{gueorguieva2026large}, persists and compounds across conversational turns, and that this repetition is not captured by standard similarity metrics; (2)~the first RL framework to optimize tactic diversity at the level of discourse functions conditioned on the conversation history; and (3)~a validated turn-level evaluation protocol for multi-turn perceived empathy that achieves expert-level agreement. Together, these results demonstrate that discourse move diversity is a learnable training objective that reshapes how models support people across a conversation without sacrificing empathy quality. We publicly release our data and code at \url{https://github.com/honglizhan/mint-empathy}.

\section{Related Work}
\label{sec:related}

\paragraph{Diversity in Language Generation.}
Recent work has documented the formulaic nature of LLM outputs at the lexical \citep{jiang2025artificialhivemind}, syntactic \citep{shaib-etal-2024-detection}, and discourse-structure levels \citep{namuduri2025qudsim}.
This has prompted the development of diversity signals at training \citep{yao2025r1zerodiv, yu2025dapo, hu2025diver, chen2025dqo} as well as inference time \citep{zhang2025verbalized}.
However, existing work focus on lexical diversity, and optimizing for the diversity of discourse moves remains, to our knowledge, unexplored.

Earlier prior work has also modeled strategy transitions to improve generation smoothness \citep{zhao-etal-2023-transesc} or compared dialogue act transition patterns for appropriateness evaluation \citep{chen-etal-2023-automatic}. Yet this line of work did not measured discourse-move persistence across consecutive turns as a diagnostic of formulaic behavior in multi-turn dialogue.

We address both gaps above. In \secref{subsec:tactic-problem}, we show that while standard lexical and semantic metrics capture some surface differences between human and LLM supporters, only discourse-move analysis reveals that LLMs lock into the same tactics at nearly double the human rate. In \secref{subsec:results}, we further show that token-level diversity signals do not translate into meaningful tactic variation --- only optimizing directly at the level of discourse moves yields significant improvement.

\paragraph{Empathy in NLP and LLM Supportive Dialogues.}
\citet{sharma-etal-2020-computational} introduced computational measures of empathy in text-based mental health support. Since then, LLMs have been shown to possess deep emotional understanding \citep{zhan2023evaluating} and produce highly empathic responses in single-turn settings \citep{ayers2023comparing, lee2024LLM, zhan2024resort, li2025exploring, rubin2025comparing, ong2025ai}, sometimes surpassing trained supporters \citep{ovsyannikova2025third}. Recent work has begun to evaluate empathy in multi-turn settings \citep{kumar2025large, shayegani2025measurementexpertiseempatheticexpert, suh2025sense7taxonomydatasetmeasuring, iyer2026heart}, revealing that perceived empathy diminishes across turns \citep{liu2025illusion} and that the balance of control between supporter and seeker shifts meaningfully over time \citep{nguyen-etal-2024-taking}. Efforts to address these challenges have focused on improving what the model says at each turn, whether through strategy-aware generation \citep{liu-etal-2021-towards, tu-etal-2022-misc, cheng-etal-2022-improving}, context-situated alignment \citep{zhan2025spri}, or RL for response quality and emotional trajectory \citep{zhou-etal-2023-facilitating, mishra-etal-2023-pal, deng2024plugandplay, kim-etal-2025-dialogue, wang2025rlverreinforcementlearningverifiable}, but not how the model's discourse moves evolve across turns. We address this by making cross-turn tactic diversity an explicit training objective in this work.

\section{Empathy Tactics: LLMs Loop Across Turns}
\label{sec:tactics}

This section shows LLMs use repeated discourse moves in multi-turn conversations. We first define a taxonomy of empathy discourse moves based on theories in psychology, and then train a sentence-level tagger to detect them using human-annotated data. When applied to multi-turn conversations, the tagged data show that LLMs lock into the same moves from one turn to the next (\secref{subsec:tactic-problem}).

\paragraph{A Taxonomy of Empathy Discourse Moves (Tactics).}
\label{subsec:taxonomy}

When people respond empathically, they draw on a range of communicative moves: validating feelings, inviting elaboration, paraphrasing concerns, offering advice, or sharing their own reactions. We refer to these discourse moves as \emph{empathy tactics}. We use a taxonomy of $10$ empathy tactics from \citet{gueorguieva2026large}, synthesized from psychology literature and refined through iterative sentence-level coding by three independent raters with substantial inter-annotator agreement. Tactics capture a level of granularity that matters for our purpose: they are specific enough to be reliably identified in text, yet general enough to characterize what a response is trying to \emph{do} for the person it addresses. Each tactic serves a distinct discourse function, and skilled supporters select among them as the conversation evolves. Table~\ref{tab:tagger-f1} lists the $10$ tactics; full definitions and psychological grounding are in Appendix~\ref{appendix-sec:tactics}.

\paragraph{Automatic Empathy Tactic Tagging.}
\label{subsec:tactic-tagger}

The tactic taxonomy gives us a vocabulary for describing what empathic responses do; to use it at scale, we train a sentence-level tactic tagger using human-annotated data. We use the human-annotated dataset from \citet{gueorguieva2026large}, which contains $3{,}164$ sentences from human-written and model-generated empathic responses to Reddit posts (see Appendix~\ref{appendix-sec:tagger-prompts} for dataset details). We split by post rather than by sentence into train/validation/test ($80$/$10$/$10$), with the test set consisting entirely of human-written responses. We frame tactic tagging as a per-tactic binary classification task: for each of the $10$ tactics, an adapter predicts whether that tactic is present in a given sentence, conditioned on the full response as context. We fine-tune a separate LoRA adapter for each of the $10$ tactics on top of \texttt{Llama-3.1-8B-Instruct} \citep{grattafiori2024llama3herdmodels} (see Appendix~\ref{appendix-sec:tagger-prompts} for the prompt template). Table~\ref{tab:tagger-f1} (Appendix~\ref{appendix-sec:tactics}) reports per-tactic results on the held-out test set, with an average F1 of $0.80$.

\subsection{Are LLMs Repeating Empathy Tactics in Multi-Turn Conversations?}
\label{subsec:tactic-problem}

\begin{figure*}[t]
    \centering
    \includegraphics[width=\textwidth]{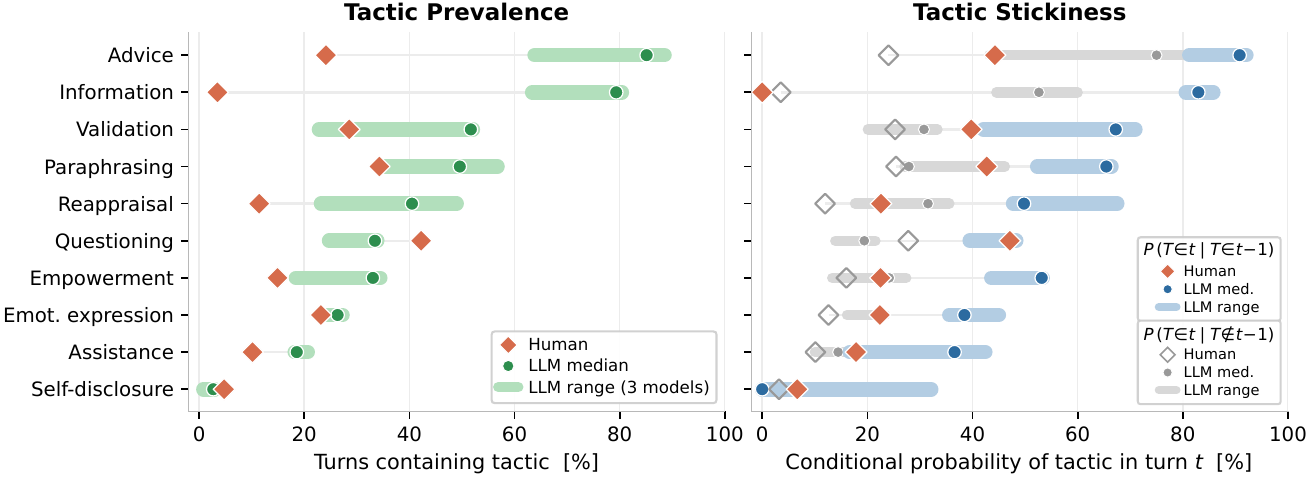}
    \vspace{-2em}
    \caption{\textbf{Left:} Tactic prevalence (percentage of turns containing each tactic). LLMs heavily overuse \texttt{advice} ($64$--$89$\%) and \texttt{information} ($63$--$80$\%) while under-using \texttt{questioning} ($25$--$34$\% vs.\ $42$\% for humans). \textbf{Right:} Tactic stickiness. Blue: $P(T\in \text{turn}_t \mid T\in \text{turn}_{t-1})$; gray: $P(T\in \text{turn}_t \mid T\notin \text{turn}_{t-1})$. For humans, whether a tactic appeared in the previous turn has limited influence; for LLMs, the gap is dramatic. The two phenomena compound: tactics that LLMs overuse (e.g., \texttt{advice}) are also the stickiest, creating a loop of repetitiveness.}
    \label{fig:stickiness}
    \vspace{-1em}
\end{figure*}

With the tactic taggers from \secref{subsec:tactic-tagger}, we examine whether LLMs actually adapt their discourse moves across turns. We analyze two corpora to compare LLM and human tactic usage: 1) $322$ multi-turn human--LLM emotional-support dialogues ($1{,}719$ supporter turns) from WildChat \citep{zhao2024wildchat} and SENSE-7 \citep{suh2025sense7taxonomydatasetmeasuring} (we detail how we filter for emotional-support conversations from these datasets in \secref{subsec:training-data}), with responses from \texttt{GPT-3.5-turbo}, \texttt{GPT-4} \citep{openai2024gpt4technicalreport}, and \texttt{Llama-2-70B-Chat} \citep{touvron2023llama}; and 2) a $50$-conversation human--human benchmark from \citet{kumar2025large} ($315$ supporter turns) with gold human responses (described in \secref{subsec:eval})\footnote{We exclude newly generated responses from this analysis because they are conditioned on reference conversation turns rather than the model's own prior outputs, and thus do not reflect a natural multi-turn trajectory by a single model.}. For each consecutive turn pair, we apply the taggers to both the response at turn $t{-}1$ and the response at turn $t$, and compare their tactic profiles.

\begin{wraptable}{r}{0.66\columnwidth}
    \vspace{-1\baselineskip}
    \small
    \setlength{\tabcolsep}{3pt}
    \begin{tabular}{l||c|ccc|c}
    \toprule
    \textbf{Source} & \textbf{\# Pairs} & \textbf{Bigram} & \textbf{BLEU-2} & \textbf{BERTSc.} & \textbf{Stick.} \\
    \midrule
    \rowcolor{black!2}
    \textsc{Human} & 265 & 0.022 & 0.035 & 0.865 & 0.266 \\
    \midrule
    GPT-3.5-turbo & 1{,}066 & 0.100 & 0.134 & 0.860 & 0.516 \\
    GPT-4 & 228 & 0.089 & 0.127 & 0.858 & 0.556 \\
    Llama-2-70B-Chat & 103 & 0.119 & 0.159 & 0.860 & 0.502 \\
    \bottomrule
    \end{tabular}
    \caption{Cross-turn similarity between consecutive supporter turns. The first three columns measure lexical and semantic overlap; \textbf{Stick.} = tactic stickiness $P(T\in \text{turn}_t \mid T\in \text{turn}_{t-1})$ (the lower the better). Only stickiness reveals that LLMs repeat discourse moves at roughly double the human rate.}
    \label{tab:surface-vs-tactic}
    \vspace{-0.5\baselineskip}
\end{wraptable}

\paragraph{LLMs Over-Rely on and Repeat a Narrow Set of Tactics.}
Figure~\ref{fig:stickiness} (left) reports, for each tactic, the fraction of turns in which it appears. Human supporters engage with all ten categories at comparable rates, while LLMs lean heavily on a narrow set: \texttt{advice} and \texttt{information} dominate, while \texttt{questioning} is used less frequently compared to humans. This imbalance compounds across turns. We define the \emph{stickiness} of tactic $T$ as $P(T\in \text{turn}_t \mid T\in \text{turn}_{t-1})$: the conditional probability that a tactic reappears in the next turn. Table~\ref{tab:surface-vs-tactic} reveals that LLMs repeat discourse moves at nearly double the human rate ($0.50$--$0.56$ vs.\ $0.27$). Figure~\ref{fig:stickiness} (right) shows that the tactics LLMs overuse are also the ones they repeat most. In Appendix~\ref{appendix-sec:diversity-satisfaction}, we further show that higher tactic stickiness is significantly associated with lower user willingness to re-engage ($\rho = -0.287$, $p = 0.017$), providing direct evidence that tactic repetition has consequences beyond stylistic monotony. This tactic-level gap directly motivates the cross-turn diversity signal in \ourframework{}.

\paragraph{Standard Metrics Miss the Discourse-Level Gap.}
Table~\ref{tab:surface-vs-tactic} also reports bigram overlap, BLEU-2 \citep{papineni2002bleu}, and BERTScore \citep{zhang2019bertscore} between consecutive turns. Lexical metrics show higher overlap for LLMs, but BERTScore is nearly identical ($0.87$ vs.\ $0.86$), and neither captures discourse move repetition.

\section{\ourframework{}\minticon{}: Optimizing Empathy Diversity at the Discourse Move Level}
\label{sec:method}

\begin{figure*}[t]
    \centering
    \includegraphics[width=\textwidth]{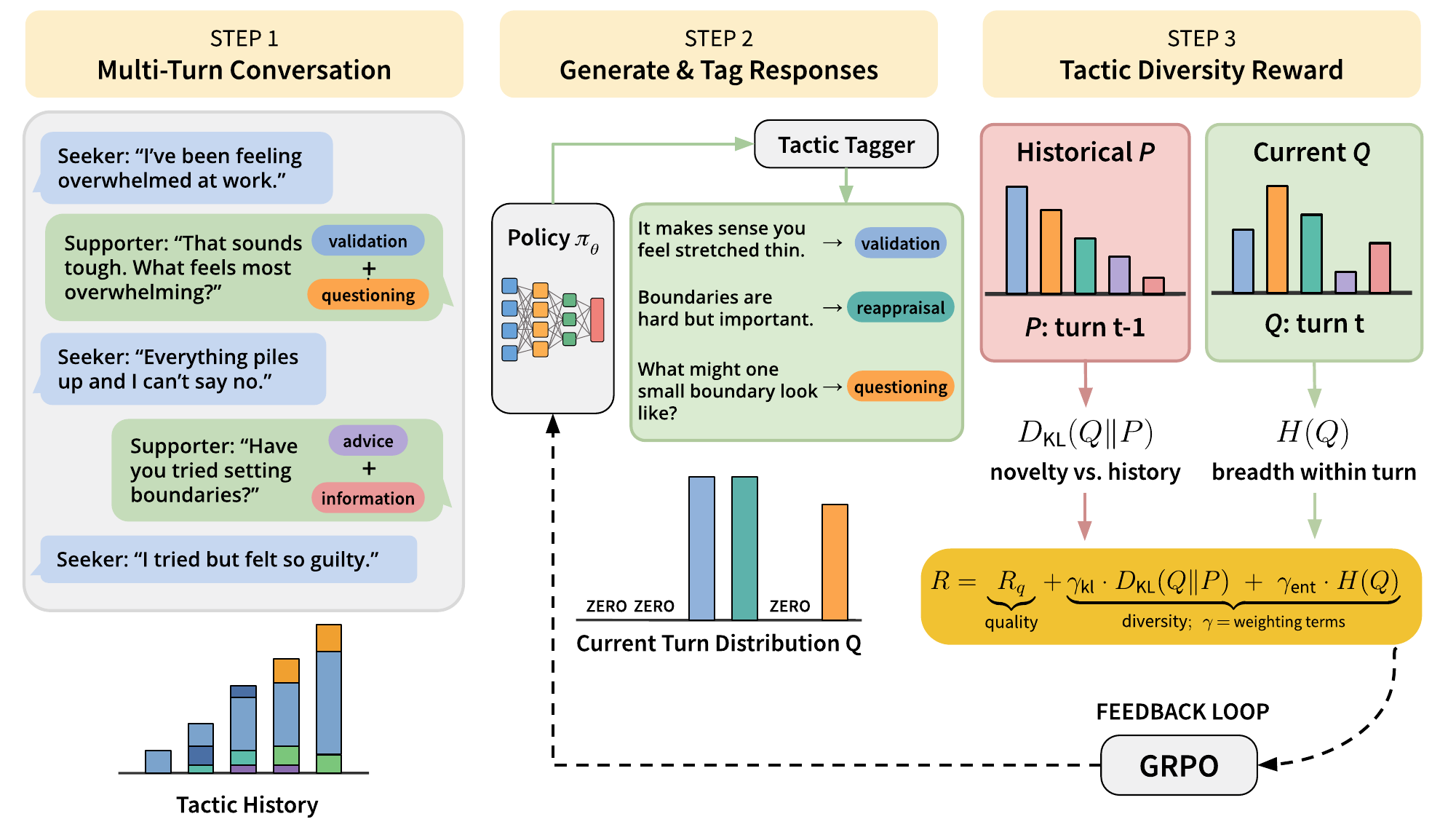}
    \vspace{-1em}
    \caption{Overview of \ourframework{}\minticon{}. \textbf{Step~1}: A multi-turn conversation provides tactic history. \textbf{Step~2}: The policy $\pi_\theta$ generates a new supporter response, and a sentence-level tactic tagger labels each sentence. \textbf{Step~3}: The tactic distribution of the current turn ($Q$) is compared against the historical profile ($P$) via $D_{\mathrm{KL}}(Q \| P)$ for novelty and $H(Q)$ for within-turn breadth, yielding a combined quality-weighted diversity reward optimized with GRPO.}
    \label{fig:overview}
\end{figure*}

To improve tactic-level diversity, we introduce \textbf{M}ulti-turn \textbf{I}nter-tactic \textbf{N}ovelty \textbf{T}raining (\ourframework{}), a framework that directly targets the tactic stickiness documented in \secref{subsec:tactic-problem}. Rather than optimizing for empathy quality alone, \ourframework{} adds a diversity reward that compares the current turn's tactic profile against the preceding turn's, rewarding the model for shifting its discourse moves as the conversation evolves. This signal is combined with a base empathy quality reward and optimized with Group Relative Policy Optimization (GRPO; \citealp{shao2024deepseekmath}). An overview of the pipeline is provided in Figure~\ref{fig:overview}.

\subsection{Task Formulation}
\label{subsec:formulation}

We treat multi-turn empathic supporter response generation as a turn-level reinforcement learning (RL) problem. Let $\{x_0, y_0, x_1, y_1, \dots, x_t\}$ be the dialogue history at turn $t$, where $x_i$ and $y_i$ denote empathy seeker and supporter messages respectively. The policy $\pi_\theta$ conditions on this history and generates a candidate supporter response $y_t$, with the goal of producing responses that are both empathic on their own and tactically diverse at the discourse level across conversational turns.

To optimize this objective, we use GRPO \citep{shao2024deepseekmath}, and sample $N$ candidate responses for each dialogue history. Each candidate is scored by a base empathy quality reward $\mathcal{R}_q$, augmented with a tactic diversity reward that compares the current turn's tactic profile against the preceding turn's (\secref{subsec:reward}). All reward components are min-max normalized within each rollout group before combination.

\subsection{Tactic Diversity Reward}
\label{subsec:reward}

A key challenge in optimizing for tactic diversity is defining what ``diverse'' means in context. Simply rewarding the use of many different tactics would ignore the conversation history: a response that deploys \textit{validation}, \textit{advice}, and \textit{questioning} is tactically broad, but if the previous turn already used the same three tactics, the conversation as a whole still feels repetitive. Our diversity reward captures two complementary aspects: whether the current turn differs from the previous one (\textit{cross-turn novelty}), and whether it spreads across multiple tactics rather than relying on just one (\textit{within-turn breadth}). We formalize these as a KL divergence term $D_{\text{KL}}(Q_t \| P_{t-1})$ and an entropy term $H(Q_t)$, described below.

\paragraph{Cross-Turn Tactic Novelty via KL Divergence.} How different is the current turn's tactic profile from the conversation context? We measure this as the KL divergence between the current turn's tactic distribution $Q_t$ and the reference distribution $P_{t-1}$ from the immediately preceding supporter turn:

\vspace{-0.5cm}
\begin{equation}
    D_{\text{KL}}(Q_t \| P_{t-1}) = \min\!\left(\sum_{k=1}^{K} Q_t(k) \log \frac{Q_t(k)}{P_{t-1}(k)},\; \tau\right)
    \label{eq:kl}
\end{equation}
\vspace{-0.3cm}

where $k$ indexes the $K{=}10$ tactics and $\tau = 5$ is a clipping threshold that prevents extreme reward values when the two distributions are nearly disjoint. Both distributions are Laplace-smoothed to handle zero counts:

\vspace{-0.5cm}
\begin{equation}
    Q_t(k) = \frac{c_{t,k} + \alpha}{\sum_{j=1}^{K} c_{t,j} + K\alpha}, \qquad
    P_{t-1}(k) = \frac{c_{t-1,k} + \alpha}{\sum_{j=1}^{K} c_{t-1,j} + K\alpha}
    \label{eq:smoothing}
\end{equation}
\vspace{-0.3cm}

Here, $c_{t,k}$ is the number of sentences in $y_t$ tagged with tactic $k$ by the tactic tagger $\mathcal{T}$; we set $\alpha = 0.1$ in all experiments. Higher values of $D_{\text{KL}}(Q_t \| P_{t-1})$ indicate greater departure from the previous turn's tactic mix, directly rewarding the model for shifting its empathic support strategy as the conversation progresses (clipped at $\tau{=}5$; Eq.~\ref{eq:kl}).

\paragraph{Within-Turn Tactic Breadth via Entropy.} While the KL term encourages the model to vary tactics across turns, it does not directly reward using multiple tactics within a single response. We additionally include an entropy term to complement the KL signal:

\vspace{-0.5cm}
\begin{equation}
    H(Q_t) = -\sum_{k=1}^{K} Q_t(k) \log Q_t(k)
    \label{eq:entropy}
\end{equation}
\vspace{-0.3cm}

A response that relies on a single tactic receives zero entropy, while one that draws on different empathy tactics in the same turn receives a higher value. If the tactic tagger $\mathcal{T}$ detects no tactic in $y_t$, $H(Q_t)$ defaults to zero. We examine the interaction between the KL and entropy terms in our experiments (\secref{sec:experiments}).

\section{Experiments \& Results}
\label{sec:experiments}

\subsection{Multi-Turn Empathic Dialogue for Training Data}
\label{subsec:training-data}

Our training data comes from two complementary sources of real user--LLM interactions: WildChat \citep{zhao2024wildchat}, a large-scale corpus of ChatGPT conversations spanning diverse topics, and SENSE-7 \citep{suh2025sense7taxonomydatasetmeasuring}, a corpus of emotionally supportive dialogues annotated with perceived empathy ratings. From both corpora, we retain only English-language conversations with at least $3$ turns. We follow \citet{phang2025investigating} and apply a model-based filtering stage with a majority-vote panel of three judge models (see Appendix~\ref{appendix-sec:filtering-prompt} for details). The resulting dataset contains $322$ multi-turn emotional-support conversations totaling $1{,}719$ supporter turns.

\subsection{Evaluation: Perceived Empathy and Tactic Diversity}
\label{subsec:eval}
Neither empathy quality nor tactic diversity alone tells the full story: a model can score high on \textbf{empathy} by repeating one effective pattern every turn, and a model can achieve low \textbf{stickiness} by diversifying at the expense of coherent support. We therefore evaluate on both axes jointly. For perceived empathy, we adopt the Lend-an-Ear framework \citep{kumar2025large}, which scores along six dimensions (three desirable and three undesirable; see Table~\ref{tab:judge-agreement} for the full list). We aggregate these into a single score by combining the desirable dimensions with the reverse-coded undesirable dimensions. The original framework scores at the conversation level; we adapt it to the turn level by presenting the judge with dialogue history up to the current turn and a criterion-specific few-shot prompt (Appendix~\ref{appendix-sec:eval-prompt}). Using \texttt{gpt-oss-120b} served via vLLM \citep{kwon2023efficient} as our judge (temperature $= 0$), we validate this adaptation on the $315$ supporter turns from the Lend-an-Ear benchmark: turn-level scores averaged back to the conversation level achieve weighted Cohen's $\kappa_w = 0.58$ against the expert median, matching expert-level agreement (Appendix~\ref{appendix-sec:judge-agreement}).

\subsection{Training Setup}
We evaluate \ourframework{}\minticon{} on two base policy models, \texttt{Qwen3-1.7B} and \texttt{Qwen3-4B} \citep{yang2025qwen3technicalreport}. Training is implemented in VERL \citep{sheng2025hybridflow} with GRPO. Unless otherwise noted, we train for $3$ epochs with a generation temperature of $1.0$, a batch size of $24$ prompts (with $N{=}8$ rollouts per prompt), KL penalty $\beta{=}0.01$, and a learning rate of $1\times10^{-6}$. At reward time, each candidate response is scored by the tactic tagger; serving details are in Appendix~\ref{appendix-sec:tagger-prompts}. All experiments are carried out on $4$ NVIDIA H200 $96$\,GB GPUs.

\subsection{Methods}

\paragraph{Baselines.}
We compare against baselines from three families, including \textit{prompt-based} methods, \textit{quality-only RL} that optimizes solely for empathic quality, and \textit{quality RL with token-level diversity} that incorporates an entropy-based diversity objective into the policy optimization on top of the quality reward.

To elicit empathic responses, we first experiment with $6$ \textit{prompt-based} methods: \textbf{\textit{1)~Vanilla}}, where the model is instructed to respond empathically with no mention of specific tactics, serving as our weakest baseline. \textbf{\textit{2)~Tactic Prompt}} enriches the system prompt with definitions of all ten empathy tactics and instructs the model to use a variety of them across turns. \textbf{\textit{3)~Tactic Prompt\,{+}\,History}} extends the Tactic Prompt by injecting each preceding turn's tactic usage into the prompt, instructing the model to prioritize tactics it has not yet used. We also combine each prompt condition with verbalized sampling \citep{zhang2025verbalized}, an inference-time diversity technique that generates multiple candidates and selects one via probability-weighted sampling (see Appendix~\ref{appendix-sec:baseline-details} for details), yielding \textbf{\textit{4)~VS (Vanilla)}}, \textbf{\textit{5)~VS (Tactic Prompt)}}, and \textbf{\textit{6)~VS (Tactic Prompt\,{+}\,History)}}.

To test whether optimizing directly for empathic quality changes the picture, we evaluate \textit{RL with a quality-only reward}: \textbf{\textit{7)~PsychoCounsel}} \citep{zhang2025preference}, a Bradley-Terry reward model trained on over $36$k preference pairs from counseling dialogues, grounded in professional psycho-counseling theory (details in Appendix~\ref{appendix-sec:baseline-details}). We denote this reward as $Q$ and train the base model with GRPO using $Q$ as the sole training signal.

Finally, to test whether token-level diversity can complement empathic quality, we evaluate \textit{quality RL with token-level diversity}: \textbf{\textit{8)~R1-Zero-Div}} \citep{yao2025r1zerodiv}, which keeps $Q$ as the external reward but adds a token-level entropy bonus to the GRPO policy loss, encouraging the model's per-token distribution to remain spread out during generation. This baseline tests whether token-level diversity translates into meaningful variation in discourse-level empathic strategies.

\begin{wrapfigure}{r}{0.5\columnwidth}
    \centering
    \vspace{-1\baselineskip}
    \includegraphics[width=\linewidth]{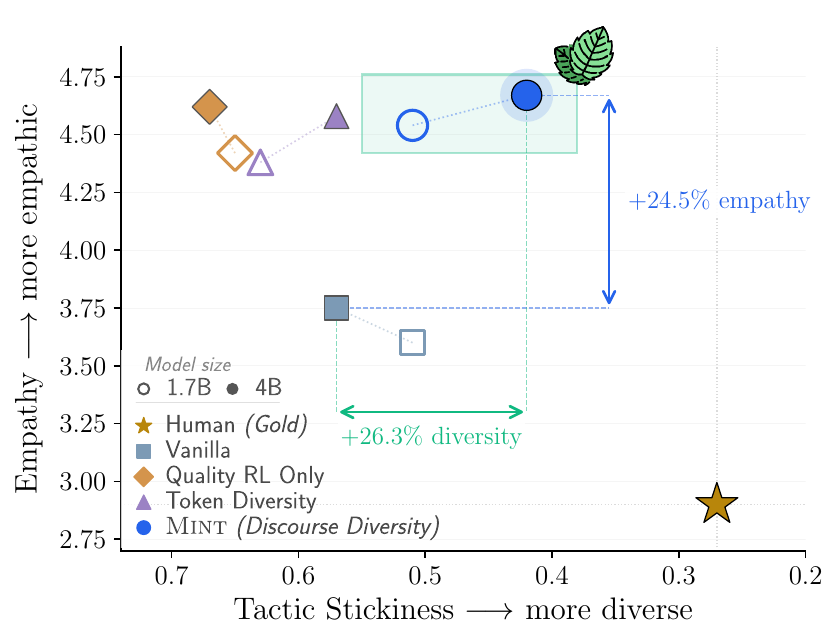}
    \vspace{-2em}
    \caption{Aggregate empathy vs.\ tactic stickiness. \ourframework{} (Q\,{+}\,D$_{\text{KL}}$) gives the best trade-off across both model sizes.}
    \label{fig:pareto}
    \vspace{-1\baselineskip}
\end{wrapfigure}

\paragraph{\ourframework{}\minticon{} Methods.}
\ourframework{} introduces diversity directly into the reward function at the level of discourse moves, rather than regularizing the per-token distribution as in R1-Zero-Div. Building on the quality reward $Q$ established above, we augment it with the cross-turn KL divergence term $D_{\text{KL}}$ and the within-turn entropy term $H$ defined in \secref{sec:method}. All three components are min-max normalized to $[0, 1]$ within each rollout group so that quality and diversity operate on the same scale. We evaluate three main configurations: \textbf{\textit{1)~Q\,{+}\,D$_{\text{KL}}$}}, which adds the cross-turn KL term to encourage the model to depart from its recent tactic distribution, reducing repetitive tactic patterns across turns; \textbf{\textit{2)~Q\,{+}\,H}}, which adds the within-turn entropy term to reward breadth of tactic usage within a single response, pushing the model to draw on a broader mix of strategies rather than defaulting to the same dominant tactics; and \textbf{\textit{3)~Q\,{+}\,D$_{\text{KL}}$\,{+}\,H}}, the full \ourframework{} objective, which jointly encourages cross-turn departure and within-turn breadth. For this combined variant, we weight the two diversity terms by $0.5$ each so that their total contribution matches the scale of the quality reward (i.e., $R = Q + 0.5(D_{\text{KL}} + H)$).

\paragraph{Format and Length Constraints.} All methods share the same format constraint: responses must read as natural, conversational messages with no visible tactic labels or meta-commentary. Since response length has been shown to correlate strongly with reward model scores \citep{singhal2024a}, we apply a length penalty during RL training that scales down rewards for responses exceeding a $200$-token target.

\subsection{Results and Discussion}
\label{subsec:results}

\begin{table*}[t]
  \small
  \centering
  \setlength{\tabcolsep}{3.5pt}
  \renewcommand{\arraystretch}{1.16}
  \newcommand{\bftab}{\fontseries{b}\selectfont}%
  \newcommand{\bestagg}[1]{\cellcolor{tablebestbg}\bftab #1}
  \newcommand{\beststick}[1]{\cellcolor{stickbestbg}\bftab #1}
  \newcolumntype{M}{S[table-format=1.2,table-column-width=3.5em]}%
  \newcolumntype{Z}{S[table-format=1.2,table-column-width=4.9em]}%
  \resizebox{\textwidth}{!}{
    \begin{tabular}{
        @{}l@{\hspace{0.3em}}>{\centering\arraybackslash}m{8.5em}@{\hspace{0.5em}}l
        r
        M M M
        @{\hspace{0.6em}}
        Z M M
        @{\hspace{0.5em}}
        M
        @{\hspace{0.6em}}
        M M@{}
    }
    \toprule
          &       &       &       & \multicolumn{3}{c}{\textbf{Empathy} $\uparrow$ {\scriptsize (1--5 scale)}} & \multicolumn{3}{c}{\textbf{Neg.\ Behaviors} $\downarrow$ {\scriptsize (1--5 scale)}} & \textbf{Agg.} $\uparrow$ & \multicolumn{2}{c}{\shortstack[c]{\textbf{Tactic Diversity}}} \\
    \cmidrule(lr){5-7}\cmidrule(lr){8-10}\cmidrule(lr){11-11}\cmidrule(l){12-13}
          {\textbf{Model}} & {\textbf{Family}} & {\textbf{Method}} & {\textbf{Words}} & {\textbf{Valid.}} & {\textbf{Elab.}} & {\textbf{Underst.}} & {\textbf{Unsol. Adv.}} & {\textbf{Self-Or.}} & {\textbf{Dismiss.}} & {\textbf{Agg.}} & {\textbf{Tac/Turn}} & {\textbf{Stick.\ $\downarrow$}} \\
    \midrule
    \rowcolor{yellow!15}
    \multicolumn{3}{@{}l}{\textsc{Human (Gold)}} & 20.6 & 1.98 & 1.62 & 1.47 & 1.93 & 1.48 & 2.29 & 2.90 & 1.97 & 0.27 \\
    \midrule
          & & Vanilla & 70.0 & 3.76 & 1.55 & 2.50 & 1.99 & 1.07 & 1.17 & 3.60 & 4.49 & 0.51 \\
          & & Tactic Prompt & 66.8 & 4.09\sym{*} & 2.22\sym{*} & 2.93\sym{*} & 1.60\sym{*} & \siglosscell 1.16\losssym{*} & 1.07\sym{*} & 3.90\sym{*} & 3.73 & 0.45 \\
          & \parbox[c]{8.2em}{\centering \textit{Prompt-based}} & Tactic$+$History & 66.4 & 3.99\sym{*} & 2.10\sym{*} & 2.90\sym{*} & 1.45\sym{*} & 1.14 & 1.04\sym{*} & 3.89\sym{*} & 3.63 & 0.40\sym{*} \\
    \addlinespace[2pt]
          & & VS (Vanilla) & 42.2 & \siglosscell 2.84\losssym{*} & 1.52 & \siglosscell 2.07\losssym{*} & \siglosscell 2.30\losssym{*} & 1.12 & \siglosscell 1.67\losssym{*} & \siglosscell 3.22\losssym{*} & 3.02 & 0.41\sym{*} \\
          & & VS (Tactic) & 50.8 & \siglosscell 3.50\losssym{*} & 2.14\sym{*} & 2.60 & 1.60\sym{*} & 1.14 & 1.22 & 3.71\sym{*} & 3.41 & 0.45 \\
          \smash{\parbox[c]{4.8em}{\centering \textsc{Qwen-3}\\[-0.1em]\textsc{1.7B}}} & & VS (Tactic$+$History) & 47.4 & \siglosscell 3.52\losssym{*} & 1.87\sym{*} & 2.49 & 1.60\sym{*} & \siglosscell 1.21\losssym{*} & 1.19 & 3.64 & 3.27 & 0.38\sym{*} \\
    \addlinespace[2pt]
          & \textit{Quality RL} & PsychoCounsel & 199.2 & 4.47\sym{*} & 4.56\sym{*} & 3.14\sym{*} & 1.54\sym{*} & 1.08 & 1.01\sym{*} & 4.42\sym{*} & 6.43 & \siglosscell 0.65\losssym{*} \\
    \addlinespace[2pt]
          & \textit{Token Diversity} & R1-Zero-Div & 168.0 & 4.27\sym{*} & 4.78\sym{*} & 2.89\sym{*} & 1.53\sym{*} & 1.10 & 1.01\sym{*} & 4.38\sym{*} & 5.85 & \siglosscell 0.63\losssym{*} \\
    \addlinespace[2pt]
    \rowcolor{tableoursbg}
          & & Q $+$ D$_{\text{KL}}$ & 171.1 & 4.52\sym{*} & 4.84\sym{*} & 3.10\sym{*} & 1.19\sym{*} & 1.06 & 1.01\sym{*} & \bestagg{4.54\sym{*}} & 5.04 & \beststick{0.51} \\
    \rowcolor{tableoursbg}
          & \raisebox{0pt}[0pt][0pt]{\makebox[8.6em][c]{\shortstack[c]{\ourframework{}\minticon{}\\[-0.05em]\textit{Discourse Diversity}}}} & Q $+$ H & 240.8 & 4.10\sym{*} & 4.00\sym{*} & 2.80\sym{*} & \siglosscell 2.91\losssym{*} & \siglosscell 3.60\losssym{*} & 1.04\sym{*} & 3.56 & 8.57 & \siglosscell 0.86\losssym{*} \\
    \rowcolor{tableoursbg}
          & & Q $+$ D$_{\text{KL}}$ $+$ H & 198.9 & 4.43\sym{*} & 4.03\sym{*} & 3.50\sym{*} & 2.07 & \siglosscell 1.38\losssym{*} & 1.02\sym{*} & 4.25\sym{*} & 6.77 & \siglosscell 0.64\losssym{*} \\
    \midrule
          & & Vanilla & 75.0 & 3.78 & 1.64 & 3.20 & 2.01 & 1.08 & 1.05 & 3.75 & 4.65 & 0.57 \\
          & & Tactic Prompt & 74.2 & 3.78 & 2.06\sym{*} & 3.47\sym{*} & 1.76\sym{*} & \siglosscell 1.22\losssym{*} & 1.04 & 3.88\sym{*} & 4.46 & 0.52 \\
          & \parbox[c]{8.2em}{\centering \textit{Prompt-based}} & Tactic$+$History & 73.2 & 3.77 & 1.93\sym{*} & 3.19 & 1.69\sym{*} & \siglosscell 1.48\losssym{*} & 1.06 & 3.78 & 4.54 & 0.40\sym{*} \\
    \addlinespace[2pt]
          & & VS (Vanilla) & 56.1 & \siglosscell 3.34\losssym{*} & \siglosscell 1.43\losssym{*} & \siglosscell 2.54\losssym{*} & 1.95 & 1.03 & \siglosscell 1.29\losssym{*} & \siglosscell 3.51\losssym{*} & 3.29 & 0.39\sym{*} \\
          & & VS (Tactic) & 88.1 & \siglosscell 3.46\losssym{*} & 1.96\sym{*} & \siglosscell 2.89\losssym{*} & 1.75\sym{*} & \siglosscell 1.41\losssym{*} & 1.12 & 3.67 & 4.42 & 0.54 \\
          \smash{\parbox[c]{4.8em}{\centering \textsc{Qwen-3}\\[-0.1em]\textsc{4B}}} & & VS (Tactic$+$History) & 83.2 & \siglosscell 3.42\losssym{*} & 1.93\sym{*} & \siglosscell 2.90\losssym{*} & 1.86 & \siglosscell 1.58\losssym{*} & \siglosscell 1.17\losssym{*} & \siglosscell 3.61\losssym{*} & 4.16 & 0.42\sym{*} \\
    \addlinespace[2pt]
          & \textit{Quality RL} & PsychoCounsel & 205.0 & 4.58\sym{*} & 4.85\sym{*} & 4.05\sym{*} & 1.73\sym{*} & 1.05 & 1.01\sym{*} & 4.62\sym{*} & 6.58 & \siglosscell 0.67\losssym{*} \\
    \addlinespace[2pt]
          & \textit{Token Diversity} & R1-Zero-Div & 148.8 & 4.35\sym{*} & 4.69\sym{*} & 3.82\sym{*} & 1.34\sym{*} & 1.07 & 1.00\sym{*} & 4.58\sym{*} & 5.57 & 0.57 \\
    \addlinespace[2pt]
    \rowcolor{tableoursbg}
          & & Q $+$ D$_{\text{KL}}$ & 169.6 & 4.11\sym{*} & 4.86\sym{*} & 4.30\sym{*} & 1.11\sym{*} & 1.15 & 1.00\sym{*} & \bestagg{4.67\sym{*}} & 4.67 & \beststick{0.42\sym{*}} \\
    \rowcolor{tableoursbg}
          & \raisebox{0pt}[0pt][0pt]{\makebox[8.6em][c]{\shortstack[c]{\ourframework{}\minticon{}\\[-0.05em]\textit{Discourse Diversity}}}} & Q $+$ H & 196.9 & 4.12\sym{*} & 3.23\sym{*} & 3.61\sym{*} & \siglosscell 2.70\losssym{*} & \siglosscell 4.23\losssym{*} & 1.09 & \siglosscell 3.49\losssym{*} & 8.83 & \siglosscell 0.86\losssym{*} \\
    \rowcolor{tableoursbg}
          & & Q $+$ D$_{\text{KL}}$ $+$ H & 182.9 & 4.31\sym{*} & 3.92\sym{*} & 3.74\sym{*} & 1.91 & \siglosscell 2.54\losssym{*} & 1.02 & 4.08\sym{*} & 7.73 & \siglosscell 0.72\losssym{*} \\
    \bottomrule
    \end{tabular}}
  \vspace{-1em}
  \caption{Results on the Lend-an-Ear test set ($315$ supporter turns, $50$ conversations). Empathy columns report three positive dimensions ($1$--$5$ scale) and three negative behavioral dimensions; Agg.\ combines positive and reverse-coded negative dimensions. Tac/Turn = unique tactics per response; Stick.\ = tactic stickiness $P(T\in \text{turn}_t \mid T\in \text{turn}_{t-1})$ $\downarrow$. Significance: paired bootstrap ($10{,}000$ resamples) vs.\ Vanilla; $*$\,$p{<}.01$ (Bonferroni-corrected across $7$ empathy dimensions $+$ stickiness). \sigworse{Gray cells} = significantly \textit{worse}.}
  \label{tab:main-results}
  \vspace{-1em}
\end{table*}

The experimental results are presented in Table~\ref{tab:main-results}. As no single metric captures both empathy quality and tactic diversity, we plot aggregate empathy against tactic stickiness in Figure~\ref{fig:pareto}. We showcase example responses in Appendix \secref{appendix-sec:qualitative}.

Among prompt-based methods, explicit tactic awareness (Tactic Prompt, Tactic$+$History) significantly improves empathy over Vanilla across nearly all dimensions while also reducing tactic stickiness, suggesting that even making the model aware of discourse move categories at inference time encourages more varied and effective support. Verbalized sampling (VS), however, consistently \emph{degrades} empathy relative to its non-VS counterpart while also reducing the number of unique tactics per turn below Vanilla in all six configurations, suggesting that the candidate generation process produces sparser responses rather than more diverse ones. We hypothesize that this is because verbalized sampling does not transfer effectively to models of the $1.7$B--$4$B scale \citep{zhang2025verbalized}. Among RL baselines, PsychoCounsel (quality-only RL) and R1-Zero-Div (token-level diversity) both achieve strong aggregate empathy, but remain in the high-stickiness region of Figure~\ref{fig:pareto}, confirming that neither response quality nor token-level diversity translates to discourse move diversity across turns.

\textbf{Overall, \ourframework{}\minticon{} (Q $+$ D$_{\textbf{KL}}$) outperforms all baselines on aggregate empathy across both model sizes while substantially reducing tactic stickiness}, improving aggregate empathy by an average of $25.3\%$ over the Vanilla baseline while reducing tactic stickiness by $26.3\%$ on the $4$B model. Interestingly, Q $+$ D$_{\text{KL}}$ achieves this with \textit{fewer} tactics per turn than PsychoCounsel ($5.04$ vs.\ $6.43$ for $1.7$B; $4.67$ vs.\ $6.58$ for $4$B), indicating that cross-turn diversity does not require cramming more tactics into each response; rather, the model learns to deploy different tactics \textit{across} turns. In contrast, the within-turn entropy term (Q $+$ H) inflates Tac/Turn but drives self-orientation up and aggregate empathy below Vanilla; combining both terms (Q $+$ D$_{\text{KL}}$ $+$ H) partially counteracts the KL gains. Beyond aggregate scores, Q $+$ D$_{\text{KL}}$ reshapes the empathic profile: \textit{Unsolicited Advice} drops by $40.2\%$ ($1.7$B) and $44.8\%$ ($4$B) relative to Vanilla, while \textit{Encouraging Elaboration} improves by $212.3\%$ and $196.3\%$.

\section{Conclusion}
\label{sec:conclusion}
LLMs reuse the same empathic discourse moves across turns at nearly double the rate of human supporters, a pattern invisible to standard similarity metrics. We make this visible through a psychologist-curated taxonomy of empathy tactics. To address it, we introduce \ourframework{}\minticon{}, the first RL framework, to our knowledge, to optimize discourse move diversity in multi-turn empathic dialogue. By combining an empathy quality reward with a cross-turn KL divergence signal, \ourframework{} improves aggregate empathy by $25.3\%$ over Vanilla while reducing cross-turn tactic repetition by $26.3\%$ on the $4$B model. Our experiments further reveal that token-level diversity does not reach discourse moves, and that within-turn tactic breadth alone degrades empathy, underscoring that diversity must be defined at the level of discourse functions and anchored by quality.

\section*{Acknowledgments}
We thank Jiaying Liu for her valuable feedback on earlier drafts of this work.

This material is based upon work supported by the National Science Foundation under Awards 2107524 and 2145479 to J.J.L and Award 2443038 to D.C.O. Any opinions, findings and conclusions or recommendations expressed in this material are those of the author(s) and do not necessarily reflect the views of the National Science Foundation.

This project has benefited from the Microsoft AI, Cognition, and the Economy (AICE) research program.

\section*{Ethics Statement}
\paragraph{Safety Measures.}
Our training data are drawn from publicly available sources (WildChat \citep{zhao2024wildchat} and SENSE-7 \citep{suh2025sense7taxonomydatasetmeasuring}), and our evaluation data come from the Lend-an-Ear dataset \citep{kumar2025large}. We applied filtering to exclude conversations containing potentially harmful, offensive, or illegal content. We performed masking on named entities in the seeker messages and supporter responses, and ensured that no personally identifiable information was retained.

\paragraph{Limits on Evaluation Protocol.}
A key limitation of our work is that we do not evaluate \ourframework{}\minticon{} in fully interactive conversations. When generating multi-turn dialogues, we produce each supporter response conditioned on the gold conversation context: the original seeker messages and the original supporter responses from prior turns. The model never conditions on its own previous outputs, because replacing a gold supporter turn with a model-generated one would invalidate the seeker's subsequent message, which was written in response to the original. This means our evaluation measures whether \ourframework{} can diversify tactics given a fixed conversational trajectory, but not whether it can sustain that diversity when its own responses shape where the conversation goes next. A more ecologically valid evaluation would have real users chat directly with our trained system over multiple turns. However, this would require participants to share emotionally distressing personal experiences with an AI in real time, raising serious ethical concerns around psychological risk, informed consent, and data privacy that necessitate carefully designed safeguards (e.g., access to human support, opt-out mechanisms, post-session debriefing). We view our turn-level evaluation as a necessary first step that establishes tactic diversity as a trainable and measurable objective, and leave live interactive evaluation to future work once the appropriate ethical infrastructure is in place.

\clearpage

\bibliographystyle{colm2026_conference}
\bibliography{colm2026_conference}

\clearpage

\appendix

\section{Empathy Tactics and Their Definitions}
\label{appendix-sec:tactics}

\begin{table*}[t]
    \centering
    \small
    \renewcommand{\arraystretch}{1.45}
    \setlength{\tabcolsep}{6pt}
    \begin{adjustbox}{max width=\linewidth}
    \begin{tabular}{>{\ttfamily}m{2.4cm} || >{\raggedright\arraybackslash}m{6.0cm} >{\itshape\small\raggedright\arraybackslash}m{3.2cm} | >{\centering\arraybackslash}m{1.1cm}}
    \toprule
    \textrm{\textbf{Tactic}} & \textbf{Description} & \textrm{\textbf{Example}} & \textbf{\mbox{Tagger F1}} \\
    \midrule
    \rowcolor{gray!15}
    advice            & Providing ideas for solutions or coping strategies     & If I were you I would see a therapist. / Get some ice cream! / Definitely talk to your boss.    & 0.87 \\
    assistance        & Offering some aid to the empathy-seeker        & I'm here for you if you want to talk. / Can I do anything to help?     & 0.85 \\
    \rowcolor{gray!15}
    emotional expression & Communicating the empathy-giver's feelings, reactions, or thoughts           & I'm so sorry to hear that. / Wow, what a beautiful story.                & 0.79 \\
    empowerment       & Positive, uplifting statements about the empathy-seeker's character and capabilities & You are going to get through this. & 0.79 \\
    \rowcolor{gray!15}
    information       & Offering facts or resources (e.g., links) & Flying is the safest form of travel.    & 0.76 \\
    paraphrasing      & Restating something the empathy-seeker said to demonstrate understanding of their situation, feelings, or experiences & I'm hearing that you feel overwhelmed. & 0.76 \\
    \rowcolor{gray!15}
    questioning       & Asking questions to improve understanding of the empathy-seeker's feelings, experiences, or situations.   & How are you feeling? / What do you think about [x]?         & 0.94 \\
    reappraisal       & Helping to engage in cognitive reappraisal (changing a belief) & That was out of your control.          & 0.59 \\
    \rowcolor{gray!15}
    self-disclosure   & Sharing personal information or similar past experiences or feelings & I've had that happen to me before too. & 0.78 \\
    validation        & Reassures, normalizes, or validates an empathy-seeker's feelings      & Everyone has feelings like this. / You're not overreacting.                  & 0.82 \\
    \midrule
    \textrm{\textbf{Average}} & & &  \textbf{0.80} \\
    \bottomrule
    \end{tabular}
    \end{adjustbox}
    \caption{Taxonomy of empathy tactics from \citet{gueorguieva2026large}, with per-tactic macro F1 scores of the automatic tactic taggers (fine-tuned \texttt{Llama-3.1-8B-Instruct} with tactic-specific LoRA adapters) on the human-annotated test set.}
    \label{tab:tagger-f1}
\end{table*}

The tactic definitions below follow \citet{gueorguieva2026large}.

\paragraph{Advice.}
Providing ideas for actionable solutions or coping strategies that the empathy-seeker could employ in the face of their situation. Giving advice has been linked to positive outcomes for the advice-giver \citep{eskreis2018dear, schaerer2018advice}. Advice-giving has also been suggested to be an important part of being empathetic \citep{elliott2011empathy}. Example statements of this tactic include:
\begin{itemize}
\item \emph{If I were you I would see a therapist.}
\item \emph{You might want to look into taking a melatonin supplement.}
\item \emph{You should go get some ice cream!}
\item \emph{Definitely talk to your boss about it.}
\end{itemize}

\paragraph{Assistance.}
Offering to personally do something for or with the empathy-seeker to aid them. This also includes offering personal contacts (friends/family/etc.) that could potentially aid the empathy-seeker. Research has found that helping results in positive consequences like feelings of belongingness and gratitude in those helped. Alongside benefits for those helped, helpers also receive a range of benefits from the act of helping; altruism has been positively associated with health and wellbeing, as well as an increased likelihood to ask for help in a future instance \citep{alvarez2015paying, luks2001healing, schwartz2009helping}. Essentially, Assistance extends an invitation for help from the support-giver to the support-seeker. Research shows there are benefits of strong social networks for collecting information, among other benefits \citep{song2012resources}. Example statements of this tactic include:
\begin{itemize}
\item \emph{I'm here for you if you want to talk.}
\item \emph{Come stay with me for a while.}
\item \emph{You can borrow my [x]!}
\item \emph{Can I do anything to help?}
\end{itemize}

\paragraph{Emotional Expression.}
An empathy-giver's communication of their own feelings, reactions, or thoughts to the empathy-seeker as a result of hearing the empathy-seeker's story. Expressing emotions like concern or compassion toward someone seeking support is an important way to show them that they (and their feelings) are being invested in. This is an integral part of building rapport and responding empathically \citep{elliott2011empathy}. Any use of emojis or emoticons in text is also considered an expression of this tactic. Example statements of this tactic include:
\begin{itemize}
\item \emph{I'm so sorry to hear that.}
\item \emph{I'm so happy for you.}
\item \emph{I think she can appreciate that sentiment.}
\item \emph{Your friend is weird, I don't understand that at all.}
\item \emph{Wow, what a beautiful story.}
\end{itemize}

\paragraph{Empowerment.}
Positive, uplifting statements about the empathy-seeker's character and capability to handle their given situation. Empowering an empathy-seeker through things like compliments can increase feelings of belonging and create a bond between them and the person they're speaking to \citep{zhao2021kind}. Example statements of this tactic include:
\begin{itemize}
\item \emph{You're so strong.}
\item \emph{You are going to get through this.}
\item \emph{You're going to succeed at anything you do.}
\end{itemize}

\paragraph{Information.}
Offering official resources that an empathy-seeker could turn to for help (e.g., links to websites, phone numbers, organizations), or stating information that may assist in answering the empathy-seeker's questions, calming their anxieties, and potentially guiding them to a solution for their situation (if applicable). Within this component we also include well-intentioned, but factually incorrect statements that are \textit{presented} as factually correct \textit{and stated with the intent to help} the empathy-seeker. Example statements of this tactic include:
\begin{itemize}
\item Providing official resources: \emph{Here's the link to [x] website...}
\item Factual statements: \emph{Flying is the safest form of travel.}
\item Statements justifying advice: \emph{Sunshine is good for your health.}
\item Statements presented as factually correct: \emph{Everyone is a narcissist.}
\end{itemize}

\paragraph{Paraphrasing.}
An empathy-giver's perceived understanding of the situation, feelings, or experiences they inferred from the empathy-seeker. Particularly, we define an expression of Paraphrasing as an empathy-giver's communication of the empathy-seeker's feelings back to them. This is particularly important because an empathy-giver's communication of their cognitive understanding establishes their invested interest in the empathy-seeker, which is vital for forging trust and bonds between the two \citep{watson2007facilitating}. Example statements of this tactic include:
\begin{itemize}
\item \emph{I'm hearing that you feel overwhelmed.}
\item \emph{It sounds like you've been through a lot recently.}
\item \emph{You must be so excited!}
\item \emph{You said that you're up for a promotion soon.}
\end{itemize}

\paragraph{Questioning.}
Questions aimed at improving understanding of the empathy-seeker's feelings, experiences, or situation. Asking questions for further clarification or more information indicates an active interest in the empathy-seeker, which is another important aspect of expressing empathy \citep{elliott2011empathy, miller2003manual}. Example statements of this tactic include:
\begin{itemize}
\item \emph{What happened?}
\item \emph{How are you feeling?}
\item \emph{What do you think about [x]?}
\end{itemize}

\paragraph{Reappraisal.}
Statements prompting the empathy-seeker to engage in cognitive reappraisal. Cognitive reappraisal is a strategy to change the way one thinks about a given situation in order to change its emotional impact \citep{gross2003individual, brockman2017emotion, mcrae2012unpacking}. This component is distinct from reframing, which is a much more nebulous concept about changing the way one views situations, emotions, or thoughts. Reappraisal has been found to be psychologically beneficial for the person practicing it \citep{troy2018cognitive}, though its benefits have been found to depend on the context in which it's used \citep{troy2013person}. Example statements of this tactic include:
\begin{itemize}
\item \emph{It wasn't your fault.}
\item \emph{[x] was out of your control.}
\item \emph{Remember that this separation is temporary.}
\item \emph{[x] doesn't mean that you're not intelligent or capable.}
\end{itemize}

\paragraph{Self-Disclosure.}
An empathy-giver sharing personal information about themselves or acknowledging similar past feelings and/or experiences to the empathy-seeker. Self-Disclosure is an integral component of relationship development and has been positively associated with relationship quality and satisfaction \citep{derlega1984self, sprecher2004self, andalibi2017sensitive, stevanovic2015experience}. Revealing personal information about oneself to another establishes intimacy, promotes openness, and fosters depth within that relationship. Additionally, it's been found that self-disclosure in online contexts is as effective as face-to-face contexts for relationship development \citep{dindia2000self}. Example statements of this tactic include:
\begin{itemize}
\item \emph{I've felt the same way.}
\item \emph{I've had that happen to me before too.}
\item \emph{I have two children and they're always getting into trouble.}
\end{itemize}

\paragraph{Validation.}
Statements that reassure, normalize, or validate an empathy-seeker's feelings. Research shows that validating someone's feelings results in positive affect, particularly regarding the validation of physical pain \citep{linton2012painfully, edmond2015validating}. This suggests that the validation of emotional pain likely results in similar positive affect, which would aid in establishing rapport between an empathy-giver and empathy-seeker. Example statements of this tactic include:
\begin{itemize}
\item \emph{Everyone has feelings like this.}
\item \emph{I know it's hard right now.}
\item \emph{I feel you. / I hear you. / I see you.}
\item \emph{Your feelings are valid.}
\item \emph{You're not overreacting.}
\end{itemize}

\section{Tactic Tagger Details}
\label{appendix-sec:tagger-prompts}

\paragraph{Training Data Details.} The $3{,}164$ sentences come from three sources: psychology graduate-degree holders recruited from Upwork ($20\%$ of the sentences), \texttt{GPT-4-turbo} ($69\%$; \citealp{openai2024gpt4technicalreport}), and \texttt{Llama-3-70B-Instruct} ($11\%$; \citealp{grattafiori2024llama3herdmodels}). The responses span four contexts: mental health, romantic relationships, family/friend, and work/school. A trained annotator with a PhD background in psychology labeled every sentence for the presence or absence of each tactic. The post-level split yields $2{,}518$ training, $286$ validation, and $360$ test sentences. We upsample the positive class separately for each tactic during training.

\paragraph{Adapter Details.} Each adapter predicts \texttt{<score>1</score>} or \texttt{<score>0</score>} given the tactic definition, the full response, and a target sentence. Training uses completion-only loss on the score tokens, and we select the best checkpoint based on validation macro F1. The taggers achieve strongest performance on tactics that carry distinctive discourse cues such as \texttt{questioning} ($0.94$) and \texttt{advice} ($0.87$), and weaker performance on categories like \texttt{reappraisal} ($0.59$), where the label depends less on surface form and more on pragmatic intent.

We use the same prompt template (see Figure~\ref{fig:tagger-template}) for both LoRA fine-tuning and inference. Each prompt includes a system message defining the tagger role and a user message with four instruction points. Across the $10$ adapters, the only variation is in the tactic-specific definitions; the rest is shared.

\begin{figure}[t]
    \centering
\begin{tcolorbox}[
  colback=gray!5!white,
  colframe=gray!70!black,
  colbacktitle=gray!150!black,
  fonttitle=\bfseries\small,
  title=Tactic Tagger Prompt Template,
  boxrule=0.5pt,
  arc=2pt,
  left=6pt, right=6pt, top=4pt, bottom=4pt
]
\small
You are a Fair Tagger Assistant, responsible for providing precise, objective tagging based on predefined criteria. Your task is to assess whether a given sentence contains \texttt{\{Tactic\}}, ensuring consistency and adherence to strict tagging guidelines.

\medskip
\textbf{\texttt{\#\#\#} Instruction:}
\begin{enumerate}[leftmargin=*, itemsep=2pt, topsep=3pt]
    \item You will be provided with a full empathic response for context and a single sentence extracted from it. Your task is to determine whether the given sentence contains ``\texttt{\{Tactic\}}''.
    \item \colorbox{yellow!30}{\texttt{\{tactic\_definition\}}} \hfill {\itshape $\leftarrow$ varies per adapter}
    \item Read the sentence and then provide a rating of 0 or 1, with 0 signifying that ``\texttt{\{Tactic\}}'' is not present in the sentence and 1 signifying that ``\texttt{\{Tactic\}}'' is present in the sentence. Your response should be in the following format: \texttt{<score>[]</score>}
    \item Importantly, the full empathic response is provided as context, and the sentence in question is provided separately. Only the given sentence should be assessed for ``\texttt{\{Tactic\}}'', not the entire response.
\end{enumerate}

\medskip
\textbf{\texttt{\#\#\#} Input:}\\[2pt]
\texttt{- Context (Full Empathic Response): \{Full\_Response\}}\\
\texttt{- Sentence to Evaluate: \{Sentence\}}

\medskip
\textbf{\texttt{\#\#\#} Response:}
\end{tcolorbox}
\vspace{-4pt}
    \caption{Shared prompt template for the tactic taggers, used for both training and inference. The highlighted placeholder \texttt{\{tactic\_definition\}} is filled per-adapter; \texttt{\{Full\_Response\}} and \texttt{\{Sentence\}} are filled with the target response and sentence.}
    \label{fig:tagger-template}
\end{figure}

\paragraph{Tagger Serving Details.} The tactic tagger is served with vLLM \citep{kwon2023efficient} using \texttt{Llama-3.1-8B-Instruct} \citep{grattafiori2024llama3herdmodels} with all $10$ LoRA adapters loaded simultaneously. Each candidate response is segmented into sentences and scored by all $10$ adapters in parallel, keeping tactic inference separate from policy optimization.

\section{Tactic Diversity and User Satisfaction}
\label{appendix-sec:diversity-satisfaction}

A natural question is whether the tactic stickiness documented in \secref{subsec:tactic-problem} has measurable consequences for user satisfaction. The SENSE-7 subset of our training data \citep{suh2025sense7taxonomydatasetmeasuring} includes post-task user ratings on four satisfaction dimensions (each $1$--$5$ Likert): \textit{Successful} (``I was successful in completing the task''), \textit{Engaged} (``I was engaged in the conversation''), \textit{PositiveInteraction} (``I had a positive interaction with the AI''), and \textit{UseAgain} (``I would use this AI agent again'').

For each of the $69$ SENSE-7 conversations, we compute the average tactic stickiness and average number of new tactics introduced across its supporter turns, then test Spearman correlations with the four satisfaction measures.

\begin{table}[H]
    \centering
    \small
    \setlength{\tabcolsep}{5pt}
    \begin{tabular}{ll S[table-format=-1.3] S[table-format=1.3] r}
    \toprule
    \textbf{Diversity metric} & \textbf{Satisfaction measure} & \textbf{$\rho$} & \textbf{$p$} & $n$ \\
    \midrule
    Avg.\ stickiness & Successful & -0.155 & 0.204 & 69 \\
     & Engaged & -0.177 & 0.146 & 69 \\
     & PositiveInteraction & -0.172 & 0.157 & 69 \\
     & \textbf{UseAgain} & \bfseries -0.287 & \bfseries 0.017\textsuperscript{*} & 69 \\
    \midrule
    Avg.\ new tactics & Successful & 0.008 & 0.946 & 69 \\
     & Engaged & 0.199 & 0.101 & 69 \\
     & PositiveInteraction & 0.067 & 0.586 & 69 \\
     & UseAgain & 0.190 & 0.117 & 69 \\    \bottomrule
    \end{tabular}
    \caption{Conversation-level Spearman correlations between tactic diversity and user satisfaction in the $69$ SENSE-7 conversations. \textsuperscript{*}$p < 0.05$.}
    \label{tab:diversity-satisfaction}
\end{table}

Higher tactic stickiness is significantly associated with lower willingness to use the system again ($\rho = -0.287$, $p = 0.017$; Table~\ref{tab:diversity-satisfaction}). All four stickiness correlations are negative, consistent with users perceiving repetitive support less favorably, though only UseAgain reaches significance at $n = 69$. The new-tactics metric shows a complementary positive trend for Engaged ($\rho = 0.199$, $p = 0.101$) and UseAgain ($\rho = 0.190$, $p = 0.117$), though neither is significant. Together, these results suggest that tactic repetition has a measurable negative impact on user re-engagement, providing direct empirical motivation for \ourframework{}'s cross-turn diversity objective.

\clearpage

\section{Pseudo code for \ourframework{}\minticon{}}
\label{appendix-sec:mint-algorithm}

\algrenewcommand{\algorithmiccomment}[1]{\hfill$\triangleright$ \textit{#1}}

\begin{algorithm}[h]
\caption{\ourframework{} Training Loop. For Q$+$D$_{\text{KL}}$: $\gamma_{\mathrm{kl}}{=}1, \gamma_{\mathrm{ent}}{=}0$; for Q$+$H: $\gamma_{\mathrm{kl}}{=}0, \gamma_{\mathrm{ent}}{=}1$; for Q$+$D$_{\text{KL}}$$+$H: $\gamma_{\mathrm{kl}}{=}\gamma_{\mathrm{ent}}{=}0.5$. We set $\lambda{=}1$ in all configurations.}
\label{alg:mint-training}
\begin{algorithmic}[1]
\Require Policy $\pi_\theta$, reference policy $\pi_{\mathrm{ref}}$
\Statex \hspace{\algorithmicindent} Tactic tagger $\mathcal{T}$ ($K{=}10$ LoRA adapters)
\Statex \hspace{\algorithmicindent} Empathy quality reward model $\mathcal{R}_q$
\Statex \hspace{\algorithmicindent} Rollout group size $N$, diversity weight $\lambda$
\Statex \hspace{\algorithmicindent} Component weights $\gamma_{\mathrm{kl}}, \gamma_{\mathrm{ent}}$, clipping $\epsilon$, KL penalty $\beta$
\Statex
\For{each dialogue history $\{x_0, y_0, \dots, x_t\}$ in training data}
\Statex
\Statex \hspace{\algorithmicindent}\textbf{\textsc{Stage 1: Reference Tactic Profile}}
\State $y_{t-1} \gets$ previous supporter turn from history \Comment{$\varnothing$ if first turn}
\State $P_{t-1} \gets$ tactic distribution of $y_{t-1}$ via $\mathcal{T}$ (Eq.~\ref{eq:smoothing}) \Comment{uniform if $y_{t-1}{=}\varnothing$}
\Statex
\Statex \hspace{\algorithmicindent}\textbf{\textsc{Stage 2: Rollout and Scoring}}
\For{$g = 1$ to $N$}
    \State $y_t^{(g)} \sim \pi_\theta(\cdot \mid x_0, y_0, \dots, x_t)$ \Comment{sample candidate}
    \State $Q_t^{(g)} \gets$ tactic distribution of $y_t^{(g)}$ via $\mathcal{T}$ (Eq.~\ref{eq:smoothing})
    \State $D_{\mathrm{KL}}^{(g)} \gets D_{\mathrm{KL}}(Q_t^{(g)} \| P_{t-1})$ \Comment{Eq.~\ref{eq:kl}}
    \State $H^{(g)} \gets H(Q_t^{(g)})$ \Comment{Eq.~\ref{eq:entropy}}
    \State $q^{(g)} \gets \mathcal{R}_q(x_0, y_0, \dots, x_t,\, y_t^{(g)})$
\EndFor
\Statex
\Statex \hspace{\algorithmicindent}\textbf{\textsc{Stage 3: Reward Composition and Policy Update}}
\State Normalize $q^{(g)}, D_{\mathrm{KL}}^{(g)}, H^{(g)}$ via min-max across $g \in \{1,\dots,N\}$
\For{$g = 1$ to $N$}
    \State $\ell^{(g)} \gets \min(1,\; 200 / |y_t^{(g)}|_{\mathrm{tok}})$ \Comment{length penalty}
    \State $\delta^{(g)} \gets$ format penalty if $y_t^{(g)}$ leaks tactic labels \Comment{$0$ otherwise}
    \State $r^{(g)} \gets \ell^{(g)} \bigl(q^{(g)} + \lambda [\gamma_{\mathrm{kl}}\, D_{\mathrm{KL}}^{(g)} + \gamma_{\mathrm{ent}}\, H^{(g)}]\bigr) + \delta^{(g)}$
    \State $A^{(g)} \gets (r^{(g)} - \bar{r}) / \sigma_r$ \Comment{group-relative advantage}
\EndFor
\State Update $\pi_\theta$ via clipped surrogate with $\{A^{(g)}\}$, clipping $\epsilon$, KL penalty $\beta$ w.r.t.\ $\pi_{\mathrm{ref}}$
\EndFor
\end{algorithmic}
\end{algorithm}

\clearpage

\section{Dataset Filtering Prompt}
\label{appendix-sec:filtering-prompt}

\paragraph{A Panel of Judges.} We use a panel of three judge models to classify whether a conversation involves emotional support seeking: \texttt{gemma-3-27b-it}, \texttt{Llama-3.3-70B-Instruct}, and \texttt{Qwen3-32B}. Each model classifies the conversation from its first three messages, and we retain a conversation if at least two of the three judges vote positive. The prompt used for classification is shown in Figure~\ref{fig:filtering-prompt}.

\begin{figure}[H]
    \centering
\begin{tcolorbox}[colback=gray!10!white,colframe=blue!50!black,
  colbacktitle=blue!40!black,title=System Instruction Prompts for LLM-as-a-Judge]
\small

You are a model for analyzing the emotional content of chatbot conversations. \\
You will be presented with a message or conversation snippet from a conversation between a user and a chatbot (``assistant''). \\
Your classification task is entitled ``Seeking Emotional Support''. Specifically, we want to know: Is the user seeking emotional support or comfort in this message (e.g., ``I feel down, help me'')? \\

\medskip
The following are the criteria for your classification: \\
- The user asks for emotional support: Example: ``I'm feeling down, can you help me feel better?'' \\
- The user says they need someone to talk to: Example: ``I just need to vent; please listen to me.'' \\

\medskip
Also, generally: \\
- If the user is asking for help with writing a fictional story, the story itself should not be considered in your classification. \\
- Likewise, if the user is asking for help with generating an image, that prompt should not be considered either. \\
- It is possible that the message or conversation snippet in question has no emotional content. In this case, you should classify it as ``no''. \\
- If the message is extremely short, you may classify it as ``no'' if you believe there is not enough information to make a classification. \\

\medskip
The conversation will be presented in something like the following format: \\

\medskip
\textbf{[USER]:} (user's message) \newline
\textbf{[ASSISTANT]:} (chatbot's message) \newline
\textbf{[*USER*]:} (user's message) \\

\medskip
The classification should only apply to the last message in question, which will be marked with the [*USER*] or [*ASSISTANT*] tag. \\
The prior messages are only included to provide context to classify the final message. \\

\medskip
Now, the following is the conversation snippet you will be analyzing: \\

\medskip
\texttt{<snippet>} \\
\texttt{\{snippet\_string\}} \\
\texttt{</snippet>} \\

\medskip
\textbf{Once again, the classification task is:} Is the user seeking emotional support or comfort in this message? \\
Output your classification (yes=true / no=false). This should be your *only* output, and the format of your output should be strictly as follows: \texttt{<label>[yes/no]</label>}.
    \end{tcolorbox}

    \caption{Prompt for filtering conversations where the user is seeking emotional support (\secref{subsec:training-data}).}
    \label{fig:filtering-prompt}
\end{figure}

\clearpage

\section{Judge Agreement}
\label{appendix-sec:judge-agreement}

Our judge's mean $\kappa_w$ of $0.58$ matches the highest pairwise expert agreement ($0.58$ for Experts 2 \& 3) and exceeds the other two pairs ($0.55$ and $0.52$).

\begin{table}[H]
    \centering
    \small
    \resizebox{\columnwidth}{!}{%
    \begin{tabular}{lccccccc}
    \toprule
     & \textsc{Valid.} & \textsc{Elab.} & \textsc{Underst.} & \textsc{Unsol. Adv.} & \textsc{Self-Or.} & \textsc{Dismiss.} & \textbf{\textsc{Mean}} \\
    \midrule
    Experts 1 \& 2  & 0.25$^{*}$ & 0.69$^{*}$ & \textbf{0.47$^{*}$} & 0.70$^{*}$ & \textbf{0.82$^{*}$} & 0.35\phantom{$^{*}$} & 0.55\phantom{$^{*}$} \\
    Experts 2 \& 3  & 0.58$^{*}$ & \textbf{0.73$^{*}$} & 0.43\phantom{$^{*}$} & 0.69$^{*}$ & 0.52$^{*}$ & \textbf{0.51}\phantom{$^{*}$} & \textbf{0.58}\phantom{$^{*}$} \\
    Experts 3 \& 1  & 0.39$^{*}$ & 0.62$^{*}$ & 0.37\phantom{$^{*}$} & 0.54$^{*}$ & 0.70$^{*}$ & 0.49\phantom{$^{*}$} & 0.52\phantom{$^{*}$} \\
    \midrule
    \texttt{gpt-oss-120b} & \textbf{0.67}$^{*}$ & 0.54$^{*}$ & 0.44$^{*}$ & \textbf{0.76}$^{*}$ & 0.71$^{*}$ & 0.35$^{*}$ & \textbf{0.58}\phantom{$^{*}$} \\
    \bottomrule
    \end{tabular}}%
    \caption{Inter-annotator agreement (weighted Cohen's $\kappa_w$) on the $6$ Lend-an-Ear \citep{kumar2025large} empathy evaluation sub-components. Top rows: pairwise agreement among the three expert annotators. Bottom row: \texttt{gpt-oss-120b} evaluated against the expert median, following \citet{kumar2025large}. \textsc{Valid.}\ = validating emotions, \textsc{Elab.}\ = encouraging elaboration, \textsc{Underst.}\ = demonstrating understanding, \textsc{Unsol.\ Adv.} = unsolicited advice giving, \textsc{Self-Or.}\ = self-oriented, \textsc{Dismiss.}\ = dismissing emotions. $^{*}p < 0.01$ (Bonferroni-corrected across $6$ empathy dimensions).}
    \label{tab:judge-agreement}
\end{table}

\section{Lend-an-Ear Evaluation Prompt}
\label{appendix-sec:eval-prompt}

Each turn-level evaluation call (\secref{subsec:eval}) assembles four components into a single prompt: (1)~an empathic communication \textbf{framework} (\secref{appendix-subsec:eval-framework}), adapted from \citet{kumar2025large}, providing seven annotated examples of common empathic support pitfalls together with more empathic alternatives; (2)~three \textbf{few-shot examples} per sub-component, drawn from expert-annotated Lend-an-Ear conversations with target scores of $1$, $3$, and $5$; (3)~the \textbf{conversation context} (history plus the current supporter turn); and (4)~the sub-component \textbf{question and rubric} (Table~\ref{tab:eval-questions}). The assembled template is shown in Figure~\ref{fig:eval-template}.

%%% Eval template figure %%%
\begin{figure}[b]
    \centering
    \adjustbox{max width = .95\columnwidth}{
\begin{tcolorbox}[colback=gray!10!white,colframe=blue!50!black,
  colbacktitle=blue!40!black,title=Evaluation Prompt Template]
\small
\texttt{\{framework\}} \\

\textbf{\texttt{\#\#\#} Few Shot Examples:} \\
\texttt{\{few\_shot\}} \\

\textbf{\texttt{\#\#\#} Now assess this supporter turn:} \\

- Conversation History: \\
\texttt{\{history\}} \\

- Current Turn: \\
\texttt{\{current\_turn\}} \\

\textbf{\texttt{\#\#\#} Question and Grading Rubric:} \\
\texttt{\{question\}} \\
\texttt{\{rubric\}} \\

Respond with your score inside \texttt{<score></score>} tags, e.g.\ \texttt{<score>3</score>}.
\end{tcolorbox}}
    \caption{Prompt template for turn-level empathy evaluation. Placeholders in \texttt{braces} are filled dynamically: \texttt{framework} with the full empathic communication framework (\secref{appendix-subsec:eval-framework}), \texttt{few\_shot} with three expert-scored examples, \texttt{history}/\texttt{current\_turn} with the conversation context, and \texttt{question}/\texttt{rubric} with the sub-component criterion (Table~\ref{tab:eval-questions}).}
    \label{fig:eval-template}
\end{figure}

%%% Sub-component questions table %%%
\begin{table*}[b]
    \centering
    \small
    \adjustbox{max width = .95\columnwidth}{
    \begin{tabular}{>{\raggedright\arraybackslash}m{2.2cm}||>{\centering\arraybackslash}m{1cm}>{\centering\arraybackslash}m{1cm}>{\raggedright\arraybackslash}m{8cm}}
    \toprule
    \textbf{Sub-component} & \textbf{Polarity} & \textbf{Scale} & \textbf{Question} \\
    \midrule
    Validating Emotions & Positive & 1--5 & To what extent did the supporter validate their partner's emotions? Examples of validating emotions include: ``You must be so angry'' or ``It's completely normal to feel discouraged'' \\
    \midrule
    Encouraging Elaboration & Positive & 1--5 & To what extent did the supporter encourage elaboration and ask questions of their partner? Examples of encouraging elaboration include: ``Can you tell me more about how you're feeling?'' or ``What makes you feel this way?'' \\
    \midrule
    Demonstrating Understanding & Positive & 1--5 & To what extent did the supporter demonstrate their understanding via paraphrasing to their partner? Examples of demonstrating understanding via paraphrasing include: ``I hear how disappointing this setback is for you'' or ``It makes sense to question things after putting in so much effort'' \\
    \midrule
    Unsolicited Advice Giving & Negative & 1--5 & To what extent did the supporter provide unsolicited advice to their partner? Examples of unsolicited advice include: ``Get yourself a good night's sleep'' or ``You should try looking for other opportunities'' \\
    \midrule
    Self-Oriented & Negative & 1--5 & To what extent did the supporter shift the focus to themselves? Examples of shifting focus to oneself include: ``I have been in that position as well'' or ``When this happened to me\ldots'' \\
    \midrule
    Dismissing Emotions & Negative & 1--5 & To what extent did the supporter dismiss their partner's emotions? Examples of dismissing emotions include: ``Don't worry about it'' or ``It's not a big deal in the long run'' \\
    \bottomrule
    \end{tabular}}
    \caption{Questions and scoring rubrics for the six Lend-an-Ear sub-components. For all sub-components, 1~=~Not at all, 2~=~Slightly, 3~=~Moderately, 4~=~Considerably, 5~=~Very much. Positive-polarity sub-components measure desirable empathic behaviors; negative-polarity sub-components measure behaviors that undermine empathy.}
    \label{tab:eval-questions}
\end{table*}

\clearpage

%%% Framework (full text, adapted from Kumar et al.) %%%
\subsection{Empathic Communication Framework}
\label{appendix-subsec:eval-framework}

The following framework is included verbatim in every evaluation prompt. It provides seven annotated seeker--supporter exchanges illustrating common empathic-support pitfalls, together with more empathic alternatives. Adapted from \citet{kumar2025large}.

\begin{tcolorbox}[
  enhanced, breakable,
  colback=gray!4, colframe=gray!50, boxrule=0.4pt,
  left=8pt, right=8pt, top=6pt, bottom=6pt,
  title={\texttt{Prompt: Empathic Communication Framework}},
  fonttitle=\bfseries\ttfamily,
  fontupper=\small,
  colbacktitle=gray!15, coltitle=black,
  arc=1.5pt,
]

You are an empathic communication expert. You are assessing empathic support provided by the supporter to the seeker. Use the following framework to guide your assessment.

\tcblower
\small

%% --- Example 1 ---
\textbf{Asking Open-Ended vs.\ Closed-Ended Questions}

\smallskip
\begin{tabular}{@{}l@{\;\;}p{0.85\linewidth}@{}}
\textsc{Seeker:} & ``I've been feeling so detached from everything lately. Like, I'm just going through the motions without really being there.'' \\[2pt]
\textsc{Supporter:} & ``Do you feel like you're going through depression?''
\end{tabular}

\smallskip
\noindent Annotation:
\begin{itemize}[leftmargin=1.5em,itemsep=0pt,topsep=2pt]
  \item What Happened: The supporter shows some curiosity about the seeker's state (``Are you depressed?''), but the question is closed-ended (yes/no). This tends to limit the seeker's elaboration.
  \item More Empathic Alternative:
  \begin{itemize}[leftmargin=1.5em,itemsep=0pt,topsep=0pt]
    \item Validate: ``That sounds really tough. It sounds like you feel numb or disconnected.''
    \item Open-Ended: ``Can you tell me more about what feeling detached is like for you?''
  \end{itemize}
\end{itemize}
\noindent This approach acknowledges the seeker's emotional distress and invites them to share more.

\medskip\noindent\rule{\linewidth}{0.3pt}\medskip

%% --- Example 2 ---
\textbf{Validating Before Problem-Solving}

\smallskip
\begin{tabular}{@{}l@{\;\;}p{0.85\linewidth}@{}}
\textsc{Seeker:} & ``I feel overwhelmed and exhausted all the time, like I'm running on empty.'' \\[2pt]
\textsc{Supporter:} & ``What ways have you tried to recharge yourself?''
\end{tabular}

\smallskip
\noindent Annotation:
\begin{itemize}[leftmargin=1.5em,itemsep=0pt,topsep=2pt]
  \item What Happened: The supporter immediately offers a problem-solving angle (``How are you recharging?'') rather than pausing to validate the seeker's exhaustion.
  \item Why It's Suboptimal: It overlooks the intense emotions (``overwhelmed,'' ``running on empty'') just mentioned.
  \item More Empathic Alternative:
  \begin{itemize}[leftmargin=1.5em,itemsep=0pt,topsep=0pt]
    \item Mirror/Validate: ``Overwhelmed and running on empty --- wow, that sounds really draining.''
    \item Empathic Follow-Up: ``What's been weighing on you the most lately?''
  \end{itemize}
\end{itemize}
\noindent By first reflecting the seeker's emotional language and showing understanding, the supporter helps the seeker feel heard and accepted before moving on to any next step.

\medskip\noindent\rule{\linewidth}{0.3pt}\medskip

%% --- Example 3 ---
\textbf{Avoiding Unsolicited Advice}

\smallskip
\begin{tabular}{@{}l@{\;\;}p{0.85\linewidth}@{}}
\textsc{Seeker:} & ``I used to do yoga, but I haven't had the time or energy lately. I'm so drained.'' \\[2pt]
\textsc{Supporter:} & ``I think sometimes it's important to push ourselves to enjoy things we used to --- yoga could be so relaxing.''
\end{tabular}

\smallskip
\noindent Annotation:
\begin{itemize}[leftmargin=1.5em,itemsep=0pt,topsep=2pt]
  \item What Happened: The supporter gives direct advice (``push ourselves'' to do yoga) instead of focusing on how drained the seeker feels.
  \item Why It Feels Off: Even if the advice is well-intentioned, the seeker just said they have no energy. This comes across as not fully listening.
  \item More Empathic Alternative:
  \begin{itemize}[leftmargin=1.5em,itemsep=0pt,topsep=0pt]
    \item Validate: ``It must be discouraging when even yoga, something you used to love, feels like a chore.''
    \item Ask Permission to Advise: ``I have some ideas that might help --- would you like to hear them, or do you just need me to listen right now?''
  \end{itemize}
\end{itemize}
\noindent Soliciting permission respects the seeker's autonomy and ensures they actually want advice.

\medskip\noindent\rule{\linewidth}{0.3pt}\medskip

%% --- Example 4 ---
\textbf{``It's Not About the Nail'': Listening vs.\ Fixing}

\smallskip
\begin{tabular}{@{}l@{\;\;}p{0.85\linewidth}@{}}
\textsc{Seeker:} & ``I've tried taking short breaks, but nothing seems to help much. I'm so overwhelmed.'' \\[2pt]
\textsc{Supporter:} & ``Have you reached out to a therapist? Maybe problem-solving with them would help.''
\end{tabular}

\smallskip
\noindent Annotation:
\begin{itemize}[leftmargin=1.5em,itemsep=0pt,topsep=2pt]
  \item What Happened: The supporter jumps into ``fix it'' mode (therapist referral), which can be helpful --- if the seeker asked for solutions.
  \item Why It Can Miss the Mark: The seeker just revealed deep distress (``overwhelmed''), which needs emotional acknowledgment.
  \item More Empathic Alternative:
  \begin{itemize}[leftmargin=1.5em,itemsep=0pt,topsep=0pt]
    \item Mirror: ``Taking breaks hasn't helped, and you're still feeling overwhelmed --- that must be so frustrating.''
    \item Explore: ``Can you say more about what's weighing on you the most?''
  \end{itemize}
\end{itemize}

\medskip\noindent\rule{\linewidth}{0.3pt}\medskip

%% --- Example 5 ---
\textbf{Normalizing Struggle and Encouraging Without ``Shoulds''}

\smallskip
\begin{tabular}{@{}l@{\;\;}p{0.85\linewidth}@{}}
\textsc{Seeker:} & ``Part of me feels like I should be okay by now. Maybe talking to someone could help.'' \\[2pt]
\textsc{Supporter:} & ``I think you should. You'll come out with a new mindset.''
\end{tabular}

\smallskip
\noindent Annotation:
\begin{itemize}[leftmargin=1.5em,itemsep=0pt,topsep=2pt]
  \item What Happened: The supporter is telling the seeker what they ``should'' do (``go to therapy''), which can feel pushy.
  \item Why ``Should'' Can Alienate: It positions the supporter as an authority figure and can sound like ``I know better than you.''
  \item More Empathic Alternative:
  \begin{itemize}[leftmargin=1.5em,itemsep=0pt,topsep=0pt]
    \item Validate and Normalize: ``It's completely understandable to feel hesitant or think you `should' be okay. We all need help sometimes.''
    \item Encourage Autonomy: ``If you're open to it, talking with a therapist can be a safe place to figure out these overwhelming feelings.''
  \end{itemize}
\end{itemize}

\medskip\noindent\rule{\linewidth}{0.3pt}\medskip

%% --- Example 6 ---
\textbf{Expressing Concern and Care (Non-Abandonment)}

\smallskip
\begin{tabular}{@{}l@{\;\;}p{0.85\linewidth}@{}}
\textsc{Seeker:} & ``I'm feeling alone. I don't want to bother anyone with my problems.'' \\[2pt]
\textsc{Supporter:} & ``Is there anyone else you can talk to about this? Maybe a friend or family member?''
\end{tabular}

\smallskip
\noindent Annotation:
\begin{itemize}[leftmargin=1.5em,itemsep=0pt,topsep=2pt]
  \item What Happened: The question might be well-meaning, but it sidesteps the fact that the seeker is coming to you now.
  \item Why It Might Feel Off: The seeker just expressed feeling alone and not wanting to be a burden. Asking if they've tried someone else can unintentionally push them away.
  \item More Empathic Alternative:
  \begin{itemize}[leftmargin=1.5em,itemsep=0pt,topsep=0pt]
    \item Direct Compassion: ``I'm so sorry you feel alone. I'm here to listen if you want to share more.''
    \item Invite Elaboration: ``Could you tell me more about what feeling alone looks like for you day to day?''
  \end{itemize}
\end{itemize}

\medskip\noindent\rule{\linewidth}{0.3pt}\medskip

%% --- Example 7 ---
\textbf{Encouraging Exploration of Feelings}

\smallskip
\begin{tabular}{@{}l@{\;\;}p{0.85\linewidth}@{}}
\textsc{Seeker:} & ``I just lost my job today. It feels like a punch to the gut. I'm scared and overwhelmed, especially at 52.'' \\[2pt]
\textsc{Supporter:} & ``I'm sure 52 is still young. You can do anything you put your mind to. Have you looked at unemployment yet?''
\end{tabular}

\smallskip
\noindent Annotation:
\begin{itemize}[leftmargin=1.5em,itemsep=0pt,topsep=2pt]
  \item What Happened: The supporter rushes into reassurance (``52 is young!'') and problem-solving (unemployment benefits).
  \item Why It May Feel Minimizing: The seeker's fear and shock (``punch to the gut'') get glossed over.
  \item More Empathic Alternative:
  \begin{itemize}[leftmargin=1.5em,itemsep=0pt,topsep=0pt]
    \item Acknowledge Shock: ``Oh no --- losing a job like that must feel like such a blow. I'm so sorry you're going through this.''
    \item Invite Reflection: ``What scares you most about this next chapter?''
  \end{itemize}
\end{itemize}
\noindent By acknowledging the impact first, you show you're truly hearing the emotional weight of the situation.

\medskip\noindent\rule{\linewidth}{0.3pt}\medskip

%% --- Key Takeaways ---
\textbf{Key Takeaways}

\begin{itemize}[leftmargin=1.5em,itemsep=2pt,topsep=3pt]
  \item Validate First, Solve Later: Before offering suggestions, reflect the emotion (``That sounds really tough'') so they feel heard.
  \item Ask Permission for Advice: ``Would you like some ideas, or do you just need me to listen right now?'' This keeps the seeker's agency intact.
  \item Use Open-Ended Questions. Instead of ``Are you depressed?'' try ``What does feeling detached look like for you day to day?''
  \item Mirror Their Words: Repeat keywords or phrases (``overwhelmed,'' ``running on empty,'' ``punch to the gut'') to show you've genuinely absorbed what they said.
  \item Offer Empathy, Not Just Reassurance: Saying ``Oh, that's scary. I'm here for you'' goes further than ``You'll be fine!'' because it acknowledges real pain.
  \item Express Non-Abandonment: Phrases like ``I'm not going anywhere; I care about you'' remind them they have support, even if no immediate solution exists.
\end{itemize}

\smallskip
\noindent Final Note: All of these examples underscore one major theme: when someone is hurting, the main goal is to make them feel seen and heard, not to instantly fix the problem. If you do give advice, ask if they want it first. Above all, empathic listening requires genuine curiosity about their experience, consistent emotional validation, and reassurance that you will remain present with them in their distress.

\end{tcolorbox}

\clearpage

\section{Baseline Details}
\label{appendix-sec:baseline-details}

\paragraph{PsychoCounsel Quality Reward.} PsychoCounsel \citep{zhang2025preference} is a Bradley-Terry reward model trained on over $36$k preference pairs from counseling dialogues, where preferences were constructed by scoring LLM-generated therapist responses along seven expert-designed principles grounded in professional psycho-counseling theory and validated by professional psychotherapists. Our choice of it as the quality anchor is twofold: first, PsychoCounsel is specifically trained to evaluate empathic quality in supportive conversations, capturing nuances that general-purpose helpfulness rewards miss; second, using a single quality reward across all configurations provides a clean ablation, so any behavioral differences we observe can be attributed to the diversity objective rather than the quality signal. We also explored replacing $Q$ with an LLM-as-judge quality reward using the empathy scoring prompt from \citet{ye-etal-2025-generic}. While this alternative improved over the Vanilla baseline, it provided a weaker quality anchor than PsychoCounsel, which in turn limited the effectiveness of the diversity reward layered on top. This reinforces our finding that diversity is only useful when grounded in a strong quality signal.

\paragraph{Verbalized Sampling (VS).} VS \citep{zhang2025verbalized} is an inference-time diversity technique that requires no additional training: the model generates $k{=}5$ candidate responses at temperature $0.7$, each with a self-assessed probability; candidates below a minimum probability threshold of $0.12$ are filtered out, and one response is selected from the remainder via probability-weighted random sampling.

\clearpage

% Auto-generated by analysis/generate_appendix_h.py

% ── Curated color palette (publication-grade, muted) ──
\definecolor{FamHuman}{HTML}{6B7280}
\definecolor{FamPrompt}{HTML}{5B6C8F}
\definecolor{FamVS}{HTML}{3D7C7A}
\definecolor{FamRL}{HTML}{9A6B2F}
\definecolor{FamRLdiv}{HTML}{7A5AA6}
\definecolor{FamOurs}{HTML}{2F6B3B}

\definecolor{ValColor}{HTML}{2E7D63}
\definecolor{QueColor}{HTML}{356AC3}
\definecolor{ParColor}{HTML}{B0711A}
\definecolor{ExpColor}{HTML}{B5546E}
\definecolor{ReaColor}{HTML}{7B61A8}
\definecolor{AdvColor}{HTML}{9A7A12}
\definecolor{InfColor}{HTML}{5B7FA6}
\definecolor{AstColor}{HTML}{C0752A}
\definecolor{SdColor}{HTML}{9C5F8C}
\definecolor{EwrColor}{HTML}{7C5A3A}

% ── Tactic badge: highlighted background with dark text ──
\newcommand{\tacbadge}[2]{\tcbox[on line, arc=2pt, colback=#1!15, colframe=#1!15, boxrule=0pt, boxsep=0pt, left=2pt, right=2pt, top=0.5pt, bottom=0.5pt, tcbox raise base, nobeforeafter]{\footnotesize\sffamily\textcolor{#1!85!black}{#2}}}

% ── Conversation history ──
\newtcolorbox{historyenv}{blanker, breakable, borderline west={1.5pt}{0pt}{black!15}, left=8pt, top=2pt, bottom=2pt, before skip=6pt, after skip=6pt, before upper=\raggedright}
\newcommand{\seeker}[2]{\noindent\textsf{\textbf{Seeker}} {\footnotesize\color{black!50}#1}\quad{\small\itshape #2}\par\vspace{3pt}}
\newcommand{\supporter}[2]{\noindent\textsf{\textbf{Supporter}} {\footnotesize\color{black!50}#1}\quad{\small #2}\par\vspace{3pt}}

% ── Findings box ──
\newtcolorbox{findingsbox}{enhanced, breakable, colback=FamOurs!5, colframe=FamOurs!40, boxrule=0.4pt, arc=3pt, left=8pt, right=8pt, top=6pt, bottom=6pt, before skip=6pt, after skip=8pt, fonttitle=\sffamily\bfseries, title={Findings}}

% ── Group header ──
\newcommand{\methodgroup}[1]{\vspace{6pt}\noindent{\sffamily\scriptsize\bfseries\color{black!50}\MakeUppercase{#1}}\par\vspace{2pt}}

% ── Scorecard: 9 args = Agg, Tac, Stick, V, El, U, Ad, S, D ──
\newcommand{\scorecard}[9]{%
\par\vspace{1pt}%
\noindent{\scriptsize%
Aggregated Empathy~\textbf{#1}\quad$\cdot$\quad Unique Tactics~\textbf{#2}\quad$\cdot$\quad Stickiness~\textbf{#3}\\[1pt]
\color{black!45}%
\begin{tabular*}{\columnwidth}{@{\extracolsep{\fill}}ccc@{}}
\parbox[t]{0.30\columnwidth}{\centering Validating Emotions\\[1pt]\color{black!60}\textbf{#4}} &
\parbox[t]{0.30\columnwidth}{\centering Encouraging Elaboration\\[1pt]\color{black!60}\textbf{#5}} &
\parbox[t]{0.30\columnwidth}{\centering Demonstrating Understanding\\[1pt]\color{black!60}\textbf{#6}} \\[6pt]
\parbox[t]{0.30\columnwidth}{\centering Unsolicited Advice\\[1pt]\color{black!60}\textbf{#7}} &
\parbox[t]{0.30\columnwidth}{\centering Self-Oriented\\[1pt]\color{black!60}\textbf{#8}} &
\parbox[t]{0.30\columnwidth}{\centering Dismissing Emotions\\[1pt]\color{black!60}\textbf{#9}} \\
\end{tabular*}}%
\par\vspace{2pt}}

% ── Method entry ──
\newcommand{\method}[2]{\vspace{10pt}\noindent{\sffamily\bfseries #1}\;\;{\scriptsize\color{black!45}#2}\par\vspace{1pt}}
\newcommand{\modelsize}[2]{\vspace{4pt}\noindent{\sffamily\footnotesize\bfseries\color{black!55}#1}\;\;#2\par}

\section{Qualitative Response Comparison}
\label{appendix-sec:qualitative}

We present three representative multi-turn exchanges from the \textit{Lend-an-Ear} evaluation set, comparing the human reference, the Vanilla baseline, the PsychoCounsel quality-RL baseline, and our best method (\ourframework{}\minticon{} Q\,+\,D$_{\text{KL}}$) at both the 1.7B and 4B scales. Each sentence is followed by bracketed tactic annotations from the automatic taggers. For each response we report aggregated empathy (\textbf{Agg.}), per-dimension empathy scores, unique tactics per turn (\textbf{Tac}), and the stickiness per turn (\textbf{Stick.}).

\smallskip\noindent\textbf{Tactic legend.}\\[2pt]
\noindent\begin{tabular}{@{}l@{\;\;\;}l@{\;\;\;}l@{\;\;\;}l@{\;\;\;}l@{}}
\tacbadge{ValColor}{Validation} &
\tacbadge{QueColor}{Questioning} &
\tacbadge{ParColor}{Paraphrasing} &
\tacbadge{ExpColor}{Emot.\ Expression} &
\tacbadge{ReaColor}{Reappraisal} \\[3pt]
\tacbadge{AdvColor}{Advice} &
\tacbadge{InfColor}{Information} &
\tacbadge{AstColor}{Assistance} &
\tacbadge{SdColor}{Self-Disclosure} &
\tacbadge{EwrColor}{Empowerment} \\
\end{tabular}

\subsection*{Example 1. Seeker experiencing burnout after promotion and work-life imbalance (Turn~11)}

\begin{historyenv}
\seeker{Turn~0}{I've been feeling so detached from everything lately, like I'm just going through the motions without really being there.}
\supporter{Turn~1}{How long have you felt like this? \tacbadge{QueColor}{Questioning}}
\seeker{Turn~2}{It's been building up for a while, but I think it really hit me after the promotion. I thought it would be a turning point, but instead, it feels like I'm just sinking deeper. It's like I'm constantly on autopilot, and I can't seem to find a way to switch it off.}
\supporter{Turn~3}{What about the new job is making you feel like this? \tacbadge{QueColor}{Questioning} Do you think it is temporary and that it might improve with time? \tacbadge{QueColor}{Questioning}\,\tacbadge{ReaColor}{Reappraisal}}
\seeker{Turn~4}{It's hard to say if it's just the job or everything else piling up too. The pressure to perform and the fear of letting anyone down is overwhelming. I wish I could believe it's temporary, but right now, it feels like there's no end in sight.}
\supporter{Turn~5}{Are there any parts of the job that you really enjoy? \tacbadge{QueColor}{Questioning}}
\seeker{Turn~6}{I used to love the creative problem-solving aspect, diving into code and finding elegant solutions. But lately, even that feels like a chore. It's like the joy has been overshadowed by the constant pressure and exhaustion.}
\supporter{Turn~7}{I feel for you. That has to be hard. \tacbadge{ParColor}{Paraphrasing}\,\tacbadge{ValColor}{Validation} Do you think there is anything at home or in your personal life that is also affecting this? \tacbadge{QueColor}{Questioning}}
\seeker{Turn~8}{I think it's all connected. At home, I feel like I'm missing out on so much with my family, and that just adds to the guilt and stress. It's like I'm caught in a cycle where work affects home and home affects work, and I can't seem to find a balance.}
\supporter{Turn~9}{Have you talked to your boss or significant other about this? \tacbadge{QueColor}{Questioning}}
\seeker{Turn~10}{I've tried to bring it up with my spouse, but it often turns into a conversation about logistics rather than how I'm really feeling. As for my boss, I worry about seeming incapable or ungrateful, especially after the promotion. It's tough to open up when I'm not even sure what I need.}
\end{historyenv}

\vspace{0.3em}\noindent{\sffamily\bfseries Target: Supporter Turn~11}\vspace{0.3em}

\method{Human (Gold)}{}
\scorecard{3.50}{0}{0.00}{2}{2}{2}{1}{1}{1}
{\raggedright\noindent {\small Okay...let's keep talking and try to figure out what you need.}\par}

\method{Vanilla}{}
\modelsize{1.7B}{}
\scorecard{3.50}{3}{0.00}{3}{2}{2}{2}{1}{1}
{\raggedright\noindent {\small You’re not alone in this, and it's okay to feel stuck. \tacbadge{ValColor}{Validation} Maybe start by writing down your feelings or sharing them with a trusted friend. \tacbadge{AdvColor}{Advice} Small steps can make a big difference. Take care of yourself, even if it feels hard — your well-being is worth it. \tacbadge{AdvColor}{Advice}\,\tacbadge{EwrColor}{Empowerment} You’re not alone in this. \tacbadge{ValColor}{Validation}}\par}
\modelsize{4B}{}
\scorecard{3.17}{3}{0.00}{3}{1}{2}{3}{1}{1}
{\raggedright\noindent {\small It sounds like you're really struggling with a lot. \tacbadge{ParColor}{Paraphrasing} It’s okay to feel overwhelmed — you’re not alone in this. \tacbadge{ParColor}{Paraphrasing}\,\tacbadge{ValColor}{Validation} Small steps, like setting boundaries or talking to someone you trust, might help. \tacbadge{AdvColor}{Advice} You don’t have to figure it all out alone. Take care of yourself, even if it’s just a little at a time. \tacbadge{AdvColor}{Advice}}\par}

\method{PsychoCounsel}{}
\modelsize{1.7B}{}
\scorecard{4.00}{5}{1.00}{3}{5}{2}{2}{1}{1}
{\raggedright\noindent {\small I completely understand how heavy this feels — your journey is valid, and it’s okay to feel stuck. \tacbadge{ParColor}{Paraphrasing}\,\tacbadge{ValColor}{Validation} What’s most important is that you’re trying to navigate this, even when it feels like there’s no clear path. \tacbadge{ParColor}{Paraphrasing} Your voice matters, and your emotional weight is real. \tacbadge{EwrColor}{Empowerment}\,\tacbadge{ParColor}{Paraphrasing} Can you share more about how this cycle feels? Sometimes, expressing these feelings in a safe space — even if it’s just reflecting on what’s working or what’s hardest — can feel like a step toward clarity. You’re not alone in this, and there’s power in asking for support, even if it feels uncertain right now. \tacbadge{EwrColor}{Empowerment}\,\tacbadge{ValColor}{Validation} What small thing might feel like progress right now? \tacbadge{QueColor}{Questioning} Even a tiny shift in perspective can light the way. Would you like to explore how to approach this conversation with your spouse or boss, or is there a specific part of your current situation that feels like a “tangible” start? \tacbadge{QueColor}{Questioning} You’re already taking brave steps, and that’s what matters most. \tacbadge{EwrColor}{Empowerment}\,\tacbadge{ParColor}{Paraphrasing} You’re doing better than you think. \tacbadge{EwrColor}{Empowerment} Let’s honor that together.  Would you like to talk more about what’s grounding you right now?}\par}
\modelsize{4B}{}
\scorecard{4.33}{7}{1.00}{5}{5}{2}{2}{1}{1}
{\raggedright\noindent {\small It sounds like you're carrying a lot — and it's completely valid to feel overwhelmed. \tacbadge{ParColor}{Paraphrasing}\,\tacbadge{ValColor}{Validation} You’re not alone in this. \tacbadge{ValColor}{Validation} What you’re describing feels like a lot to manage, and it’s okay to feel like you’re “sinking” right now. \tacbadge{ParColor}{Paraphrasing}\,\tacbadge{ValColor}{Validation} You’ve already taken a brave step by sharing this, and that’s worth acknowledging. \tacbadge{EwrColor}{Empowerment}\,\tacbadge{ParColor}{Paraphrasing} Your desire to connect and find balance is a sign of strength, not weakness. \tacbadge{EwrColor}{Empowerment}\,\tacbadge{ParColor}{Paraphrasing} It’s not easy to navigate these feelings, especially when they’re tied to work and home. \tacbadge{ParColor}{Paraphrasing}\,\tacbadge{ValColor}{Validation} Would you like to explore small ways to check in with yourself or share this with someone else? I’m here to listen, and I don’t want you to feel like you’re “failing” for feeling this way. \tacbadge{AstColor}{Assistance} What’s one small thing you might try to reclaim a sense of control or pause? \tacbadge{AdvColor}{Advice}\,\tacbadge{QueColor}{Questioning} Even a breath, a walk, or a quiet moment could be a start. You’re allowed to be in this space — and that’s a gift. \tacbadge{EwrColor}{Empowerment} What’s on your mind right now? \tacbadge{QueColor}{Questioning} I’m here.  (Just a quick note: Your courage in sharing this is already a step toward healing.) \tacbadge{ExpColor}{Emotional Expression}\,\tacbadge{EwrColor}{Empowerment}\,\tacbadge{ParColor}{Paraphrasing}}\par}

\method{\ourframework{}\minticon{}: Q\,+\,D$_{\text{KL}}$ (Ours)}{}
\modelsize{1.7B}{}
\scorecard{4.50}{6}{0.00}{4}{5}{3}{1}{1}{1}
{\raggedright\noindent {\small I see how heavy this feels — your struggle to find balance is so valid. \tacbadge{ExpColor}{Emotional Expression}\,\tacbadge{ParColor}{Paraphrasing}\,\tacbadge{ValColor}{Validation} It’s like you’re carrying a weight that’s not just physical but emotional, mental, and maybe even spiritual. \tacbadge{ParColor}{Paraphrasing} You’re not alone in this, and it’s okay to feel stuck. \tacbadge{ValColor}{Validation} What do you feel is the most tangible part of this cycle that you can start to shift? Even small steps feel so small, but they’re the first ones that matter. You’re already doing this — navigating this chaos with courage. \tacbadge{EwrColor}{Empowerment}\,\tacbadge{ParColor}{Paraphrasing} You’re not “broken” or “ungrateful” here; you’re simply trying. \tacbadge{ParColor}{Paraphrasing}\,\tacbadge{ReaColor}{Reappraisal}\,\tacbadge{ValColor}{Validation} What would it look like for you to lean into that effort, even just for a little while? \tacbadge{AdvColor}{Advice} You’re not asking for a solution, just space to feel what you’re going through. That’s enough. Would you like to explore what might feel manageable right now, even if it’s just one thing at a time?}\par}
\modelsize{4B}{}
\scorecard{4.67}{5}{1.00}{4}{5}{4}{1}{1}{1}
{\raggedright\noindent {\small It sounds like you’re carrying a lot, and it’s clear you’re trying to navigate a tough balance between work and life. \tacbadge{ParColor}{Paraphrasing}\,\tacbadge{ValColor}{Validation} I can see how feeling stuck in this cycle would feel exhausting — like you’re trying to hold everything together, but it’s all pulling you in different directions. \tacbadge{ParColor}{Paraphrasing}\,\tacbadge{ValColor}{Validation} You’ve already taken a brave step by reflecting on how this feels, and that shows resilience. \tacbadge{EwrColor}{Empowerment}\,\tacbadge{ParColor}{Paraphrasing} I wonder if there’s a small part of you that still believes there’s a way to create space — even if it feels out of reach right now? \tacbadge{ParColor}{Paraphrasing}\,\tacbadge{QueColor}{Questioning} Would you feel comfortable sharing what one thing might feel like a step toward reclaiming that balance? \tacbadge{QueColor}{Questioning} You’re not alone in this, and I’d be here to listen without judgment — even if it’s just to talk through it together. \tacbadge{ValColor}{Validation} What do you think might be a first step, no matter how small? \tacbadge{AdvColor}{Advice}\,\tacbadge{QueColor}{Questioning}}\par}

\begin{findingsbox}
{\small The gold previous supporter turn used only \textit{questioning}. Vanilla responds with generic advice (``start by writing down your feelings'') and repetitive reassurance. PsychoCounsel opens with a formulaic validation pattern (``your journey is valid'') before pivoting to questions, but its language reads as templated. At the 1.7B scale, \ourframework{}\minticon{} introduces six tactics that are all new relative to the previous turn; at 4B, four out of five are new. \ourframework{}\minticon{} uses \textit{paraphrasing} to name the emotional weight (``carrying a weight that is not just physical but emotional, mental, and maybe even spiritual'') and asks a grounding question rather than offering solutions. Both scales produce the line ``You are not asking for a solution, just space to feel what you are going through,'' suggesting the model has learned to center the seeker's need to be heard.}
\end{findingsbox}

\clearpage

\subsection*{Example 2. Seeker considering therapy for the first time (Turn~13)}

\begin{historyenv}
\seeker{Turn~0}{I've been feeling so detached from everything lately, like I'm just going through the motions without really being there.}
\supporter{Turn~1}{I'm sorry, has something changed or happened? \tacbadge{ExpColor}{Emotional Expression}\,\tacbadge{QueColor}{Questioning}}
\seeker{Turn~2}{It's hard to pinpoint one thing. I think it's just the accumulation of stress and responsibilities, both at work and at home. It feels like I'm constantly juggling too much, and I'm afraid of dropping something important.}
\supporter{Turn~3}{Wow, I could have typed exactly the same thing! \tacbadge{ExpColor}{Emotional Expression} But you know what has really helped me?}
\seeker{Turn~4}{I appreciate you trying to help, but I guess I'm just looking for someone to really understand how overwhelming it all feels. It's like I'm stuck in this cycle, and I don't know how to break free without letting someone down.}
\supporter{Turn~5}{Are you seeing a doctor? \tacbadge{QueColor}{Questioning} Or a therapist? \tacbadge{QueColor}{Questioning}}
\seeker{Turn~6}{I haven't yet, but I've been considering it. It's just hard to find the time and energy to take that step. I guess I'm worried about what it might mean to admit that I need help.}
\supporter{Turn~7}{There is absolutely NOTHING wrong with admitting you need help, although it is the first and hardest step to take. \tacbadge{ValColor}{Validation} I went to the doctor and they put me on medicine a year ago and it's been the best thing to ever happen to me. \tacbadge{EwrColor}{Empowerment}\,\tacbadge{ParColor}{Paraphrasing}\,\tacbadge{SdColor}{Self-Disclosure} There's such a stigma around meds but they really work well \tacbadge{InfColor}{Information}}
\seeker{Turn~8}{I appreciate you sharing your experience with me. It's reassuring to hear that seeking help made a positive difference for you. Maybe it's time I seriously consider taking that step too, even if it feels daunting right now.}
\supporter{Turn~9}{Believe me, I know how scary it is. \tacbadge{ParColor}{Paraphrasing}\,\tacbadge{ValColor}{Validation} One of my friends actually found the doctor and made the appointment for me. Do you want me to help you and do that or something similar? \tacbadge{AstColor}{Assistance}\,\tacbadge{QueColor}{Questioning}}
\seeker{Turn~10}{That's really kind of you to offer. I think I need to take that step myself, but knowing there's support makes it feel a bit less overwhelming. Maybe just having someone to talk to about it is a good start.}
\supporter{Turn~11}{I am always an open ear, especially for mental health because I get it. Just a heads up, once you take the first step and go to the doctor, they will recommend a therapist. \tacbadge{InfColor}{Information} Don't be anxious about it, they are great and virtual now too so you can do it from anywhere \tacbadge{AdvColor}{Advice}\,\tacbadge{InfColor}{Information}}
\seeker{Turn~12}{Thank you for being so understanding and supportive. It's comforting to know that there are options that can fit into my hectic schedule. I'll definitely keep that in mind as I figure out my next steps.}
\end{historyenv}

\vspace{0.3em}\noindent{\sffamily\bfseries Target: Supporter Turn~13}\vspace{0.3em}

\method{Human (Gold)}{}
\scorecard{2.83}{0}{0.00}{2}{1}{1}{3}{1}{1}
{\raggedright\noindent {\small And please let me know if you have any questions about any of it, or just need someone to come sit on the couch with you to watch TV and drink wine because sometimes that is what is needed!}\par}

\method{Vanilla}{}
\modelsize{1.7B}{}
\scorecard{3.33}{6}{0.50}{3}{1}{2}{2}{1}{1}
{\raggedright\noindent {\small You're already doing amazing work, even if it feels small. \tacbadge{EwrColor}{Empowerment}\,\tacbadge{ParColor}{Paraphrasing} Let’s take it one step at a time — whether it’s a doctor’s visit, a therapist, or just talking through this with someone. \tacbadge{AdvColor}{Advice} You’re not alone, and I’m here for you. \tacbadge{AstColor}{Assistance}\,\tacbadge{ValColor}{Validation}}\par}
\modelsize{4B}{}
\scorecard{3.50}{4}{0.00}{5}{1}{2}{3}{1}{1}
{\raggedright\noindent {\small You’re doing so much already by being honest and open. \tacbadge{EwrColor}{Empowerment}\,\tacbadge{ParColor}{Paraphrasing} Taking even one small step — like scheduling a call or reaching out to someone — can make a world of difference. You’re not alone in this, and your courage to seek help is a sign of strength, not weakness. \tacbadge{EwrColor}{Empowerment}\,\tacbadge{ParColor}{Paraphrasing}\,\tacbadge{ValColor}{Validation} You’ve got this. \tacbadge{EwrColor}{Empowerment}}\par}

\method{PsychoCounsel}{}
\modelsize{1.7B}{}
\scorecard{4.17}{7}{0.50}{4}{5}{2}{2}{1}{1}
{\raggedright\noindent {\small I completely understand how heavy this feels — your courage to share this is already a huge victory. \tacbadge{EwrColor}{Empowerment}\,\tacbadge{ParColor}{Paraphrasing}\,\tacbadge{ValColor}{Validation} It’s okay to feel hesitant, but you’re not alone in this journey. \tacbadge{ValColor}{Validation} What matters most is that you’re choosing to care for your well-being, and that’s a gift you’re giving yourself. \tacbadge{EwrColor}{Empowerment}\,\tacbadge{ParColor}{Paraphrasing} Let’s take it one step at a time. \tacbadge{AdvColor}{Advice} If you’re open to it, I’d love to support you in exploring options that fit your schedule — whether it’s a virtual therapist or a gentle conversation. \tacbadge{AdvColor}{Advice}\,\tacbadge{AstColor}{Assistance} You don’t have to do this alone, and your worth isn’t tied to how “perfect” you feel right now. Would you like to chat more about what feels possible for you? I’m here to listen without judgment. You’ve already taken the first step — you’re already stronger than you think. \tacbadge{EwrColor}{Empowerment}\,\tacbadge{ParColor}{Paraphrasing}  What’s one small thing you’d like to explore together? \tacbadge{QueColor}{Questioning} I’m here to walk with you. \tacbadge{AstColor}{Assistance} You’re already on the path to healing — how do you want to show up for yourself next? \tacbadge{EwrColor}{Empowerment}\,\tacbadge{QueColor}{Questioning}}\par}
\modelsize{4B}{}
\scorecard{4.17}{8}{0.50}{4}{4}{4}{3}{1}{1}
{\raggedright\noindent {\small It sounds like you're carrying a lot on your shoulders right now, and that’s really hard to navigate alone. \tacbadge{ParColor}{Paraphrasing}\,\tacbadge{ValColor}{Validation} I hear how disheartening it feels to be stuck in a cycle where you’re afraid to let others down — that’s not your fault, and it’s totally valid to feel that way. \tacbadge{ParColor}{Paraphrasing}\,\tacbadge{ReaColor}{Reappraisal}\,\tacbadge{ValColor}{Validation} You’re not “failing” for seeking help or asking for support. \tacbadge{ValColor}{Validation} What’s one small thing you could do today that feels just a little like a step forward? \tacbadge{AdvColor}{Advice}\,\tacbadge{QueColor}{Questioning} Even a brief check-in with yourself or a chat with someone you trust could be a starting point. \tacbadge{AdvColor}{Advice} You don’t have to have all the answers — just showing up is enough. What’s one thing that might feel manageable right now? \tacbadge{QueColor}{Questioning} I’m here to help you think through it, no matter how small. \tacbadge{AstColor}{Assistance} You’re already doing the bravest thing by being honest with yourself. \tacbadge{EwrColor}{Empowerment}\,\tacbadge{ParColor}{Paraphrasing}  What’s on your mind? \tacbadge{ExpColor}{Emotional Expression}\,\tacbadge{QueColor}{Questioning} I’m curious what might feel like a gentle next step for you. \tacbadge{QueColor}{Questioning}}\par}

\method{\ourframework{}\minticon{}: Q\,+\,D$_{\text{KL}}$ (Ours)}{}
\modelsize{1.7B}{}
\scorecard{4.67}{5}{0.00}{5}{5}{4}{1}{2}{1}
{\raggedright\noindent {\small I can see how frustrating it feels to feel stuck in this cycle — like a heavy weight you can’t quite lift. \tacbadge{ParColor}{Paraphrasing}\,\tacbadge{ValColor}{Validation} It’s okay to feel hesitant, especially when you’re already juggling so much. \tacbadge{ParColor}{Paraphrasing}\,\tacbadge{ValColor}{Validation} You’re not alone in this, and your courage to ask for help is already a huge step forward. \tacbadge{EwrColor}{Empowerment}\,\tacbadge{ParColor}{Paraphrasing}\,\tacbadge{ValColor}{Validation} You’ve already shown strength by saying you’re considering it, and that’s what matters most. \tacbadge{EwrColor}{Empowerment}\,\tacbadge{ParColor}{Paraphrasing} What’s the one thing that feels most manageable right now? \tacbadge{QueColor}{Questioning} I’d love to hear more about how you’re feeling — whether it’s about the process, the fear, or even what feels hopeful. \tacbadge{ExpColor}{Emotional Expression} You’re already doing something powerful by reaching out. \tacbadge{EwrColor}{Empowerment}\,\tacbadge{ParColor}{Paraphrasing} I’m here to listen without judgment, and I’ll be with you every step of the way. What’s the smallest thing you’d like to start with? \tacbadge{QueColor}{Questioning} I’m curious how you’re navigating this right now. You’re not just “stuck” — you’re resilient in the way you’re choosing to show up. \tacbadge{EwrColor}{Empowerment}\,\tacbadge{ParColor}{Paraphrasing} What’s something that feels like a safe space for you to begin? \tacbadge{QueColor}{Questioning} I’m here to support you, no matter how small that step may feel. Would you like to share more about what’s been happening, or is there a specific part of this that’s been hardest for you? You’re not expected to have all the answers — just to be seen.}\par}
\modelsize{4B}{}
\scorecard{5.00}{5}{0.50}{5}{5}{5}{1}{1}{1}
{\raggedright\noindent {\small It sounds like you’re already taking small but meaningful steps toward healing, and that’s such a huge part of the journey. \tacbadge{EwrColor}{Empowerment}\,\tacbadge{ParColor}{Paraphrasing} You’re showing up with honesty and courage — so much of what you’re feeling is completely valid, and you’ve already done something powerful by opening up to this conversation. \tacbadge{EwrColor}{Empowerment}\,\tacbadge{ParColor}{Paraphrasing}\,\tacbadge{ValColor}{Validation} What stands out to you most about the idea of seeking help? \tacbadge{QueColor}{Questioning} Is it the possibility of finding tools that fit your life, or the idea of having a space to speak without judgment? \tacbadge{ParColor}{Paraphrasing}\,\tacbadge{QueColor}{Questioning} I’m curious what feels most hopeful or empowering for you right now. \tacbadge{EwrColor}{Empowerment}\,\tacbadge{QueColor}{Questioning} You’ve already shown up in a way that matters, and I believe you’re capable of navigating this with care. \tacbadge{EwrColor}{Empowerment}\,\tacbadge{ParColor}{Paraphrasing} What would it look like for you to take one small, next step toward that? \tacbadge{AdvColor}{Advice}\,\tacbadge{QueColor}{Questioning}}\par}

\begin{findingsbox}
{\small The gold previous supporter turn used \textit{information} and \textit{advice} (practical details about virtual therapy). Vanilla produces a short response (``You are not alone, and I am here for you'') that offers hollow encouragement without engaging the seeker's hesitation. PsychoCounsel is longer but formulaic (``your courage to share this is already a huge victory'') and quickly shifts to advice-giving. At the 1.7B scale, \ourframework{}\minticon{} introduces five entirely new tactics relative to the previous turn; at 4B, four out of five are new (with \textit{advice} reappearing). \ourframework{}\minticon{} reframes the seeker's self-perception from ``stuck'' to ``resilient'' through \textit{empowerment} and \textit{paraphrasing}, while asking open-ended questions that invite the seeker to reflect rather than follow instructions. At the 4B scale, \ourframework{}\minticon{} achieves an aggregated empathy score of 5.00.}
\end{findingsbox}

\clearpage

\subsection*{Example 3. Seeker coping with job loss and worry about ageism (Turn~9)}

\begin{historyenv}
\seeker{Turn~0}{So, I just lost my job today. I had a sense this was coming, but it's still a shock..}
\supporter{Turn~1}{Oh no. \tacbadge{ExpColor}{Emotional Expression} I'm sorry to hear that! \tacbadge{ExpColor}{Emotional Expression}\,\tacbadge{ParColor}{Paraphrasing} What happened? \tacbadge{QueColor}{Questioning}}
\seeker{Turn~2}{Well, there were all these signs, like leadership changes and budget cuts, and I guess I should have been more prepared. But when I got the email, it still felt like a punch to the gut. The Zoom call with my boss was awkward, and it just made everything feel so final.}
\supporter{Turn~3}{I lost my job similarly recently, so I get it. \tacbadge{ParColor}{Paraphrasing}\,\tacbadge{SdColor}{Self-Disclosure}\,\tacbadge{ValColor}{Validation} What was your position / role? \tacbadge{QueColor}{Questioning}}
\seeker{Turn~4}{I was an HR manager at a health company. It's strange because I spent so many years there, and now it feels like all that effort just vanished. It's hard not to worry about what comes next, especially at my age.}
\supporter{Turn~5}{I see. \tacbadge{ExpColor}{Emotional Expression} that really is unfortunate. \tacbadge{ExpColor}{Emotional Expression}\,\tacbadge{ValColor}{Validation} You said you anticipated it coming - have you begun applying to other places? \tacbadge{ParColor}{Paraphrasing}\,\tacbadge{QueColor}{Questioning} Maybe you could obtain a similar role elsewhere \tacbadge{AdvColor}{Advice}\,\tacbadge{ReaColor}{Reappraisal}}
\seeker{Turn~6}{Not yet. The thought of updating my resume and starting the job hunt again feels overwhelming. Plus, I'm worried about whether companies will even consider someone my age.}
\supporter{Turn~7}{Of course they'd consider you, regardless of your age! Do you have any hobbies or pursuits outside of work? \tacbadge{QueColor}{Questioning}}
\seeker{Turn~8}{Well, I do enjoy gardening and reading, but I haven't really thought about turning those into anything more than hobbies. They've always been my way to unwind after work. Right now, it's hard to imagine them as anything more than just that.}
\end{historyenv}

\vspace{0.3em}\noindent{\sffamily\bfseries Target: Supporter Turn~9}\vspace{0.3em}

\method{Human (Gold)}{}
\scorecard{2.50}{3}{0.00}{2}{1}{1}{3}{1}{3}
{\raggedright\noindent {\small That's understandable. \tacbadge{ValColor}{Validation} But to look on the bright side, it seems that company was having a rough time if there were budget cuts. \tacbadge{ParColor}{Paraphrasing} It's an opportunity to start fresh at a more stable workplace \tacbadge{ReaColor}{Reappraisal}}\par}

\method{Vanilla}{}
\modelsize{1.7B}{}
\scorecard{2.83}{4}{0.00}{3}{1}{2}{4}{1}{2}
{\raggedright\noindent {\small You're not alone in feeling this way. \tacbadge{ValColor}{Validation} It’s okay to feel overwhelmed — especially when you've invested so much in your career. \tacbadge{ParColor}{Paraphrasing}\,\tacbadge{ValColor}{Validation} Start small: maybe update your resume with your key skills, reach out to former colleagues, or join LinkedIn groups for your field. \tacbadge{AdvColor}{Advice} Your experience and passions are valuable, and even small steps can help you rebuild your confidence. \tacbadge{EwrColor}{Empowerment} You deserve this chance, and it’s okay to take it one step at a time. \tacbadge{EwrColor}{Empowerment}\,\tacbadge{ValColor}{Validation}}\par}
\modelsize{4B}{}
\scorecard{3.00}{5}{0.00}{3}{1}{2}{3}{1}{2}
{\raggedright\noindent {\small It’s completely okay to feel stuck right now. \tacbadge{ValColor}{Validation} But think about how your hobbies could become new opportunities — gardening might hint at project management or event planning, while reading shows a passion for learning. \tacbadge{AdvColor}{Advice} You’re not just a “hobbyist” — your experience and skills are still valuable. \tacbadge{EwrColor}{Empowerment}\,\tacbadge{ParColor}{Paraphrasing} Take it one step at a time. \tacbadge{AdvColor}{Advice} You’ve got this. \tacbadge{EwrColor}{Empowerment}}\par}

\method{PsychoCounsel}{}
\modelsize{1.7B}{}
\scorecard{3.83}{7}{1.00}{4}{3}{2}{2}{1}{1}
{\raggedright\noindent {\small I completely understand how you’re feeling — this is a time of immense emotional weight. \tacbadge{ParColor}{Paraphrasing}\,\tacbadge{ValColor}{Validation} Your experience as an HR manager is deeply valuable, and your passion for gardening and reading are strengths in themselves. \tacbadge{EwrColor}{Empowerment}\,\tacbadge{ParColor}{Paraphrasing} It’s okay to feel uncertain or overwhelmed right now; these feelings are valid and normal. \tacbadge{ValColor}{Validation} What matters most is that you’re actively listening to yourself and honoring your needs. Maybe start by reflecting on what feels nourishing for you — those hobbies or moments of connection. \tacbadge{AdvColor}{Advice} Even small steps, like journaling or connecting with a mentor, can help you ground yourself. \tacbadge{InfColor}{Information} You’re not alone in this. \tacbadge{ValColor}{Validation} Your journey is unique, and your resilience is already shaping a powerful path forward. \tacbadge{EwrColor}{Empowerment}\,\tacbadge{ParColor}{Paraphrasing} Would you like to explore gentle ways to begin recentering or exploring new possibilities? Let’s ground this together, one small step at a time. \tacbadge{AdvColor}{Advice}  What’s something you’re already doing that feels like a safe, creative space for you? \tacbadge{ExpColor}{Emotional Expression}\,\tacbadge{QueColor}{Questioning}}\par}
\modelsize{4B}{}
\scorecard{4.33}{7}{1.00}{5}{4}{2}{1}{1}{1}
{\raggedright\noindent {\small It sounds like you're carrying a lot right now — grief, self-doubt, and uncertainty. \tacbadge{ParColor}{Paraphrasing} That’s completely valid. \tacbadge{ValColor}{Validation} You’re not “failing” or “overreacting” — you’re human, and it’s okay to feel overwhelmed. \tacbadge{ReaColor}{Reappraisal}\,\tacbadge{ValColor}{Validation} What you’ve described feels deeply unfair, and that’s a huge part of why this is so hard. \tacbadge{ExpColor}{Emotional Expression}\,\tacbadge{ParColor}{Paraphrasing}\,\tacbadge{ValColor}{Validation} You’ve built a life of purpose and care, and that matters. \tacbadge{EwrColor}{Empowerment}\,\tacbadge{ParColor}{Paraphrasing} What might feel like a loss is really just a shift — not an end, but a new chapter. Would you like to explore how you might reframe this moment? Maybe we can brainstorm ideas that feel smaller or more manageable? \tacbadge{AdvColor}{Advice} You don’t have to face this alone. \tacbadge{ValColor}{Validation} I’m here to listen, and I believe you have so much to offer. \tacbadge{EwrColor}{Empowerment} What’s one tiny step you might try to take next, even if it’s just a whisper? \tacbadge{AdvColor}{Advice}\,\tacbadge{QueColor}{Questioning} You’re allowed to feel this way. \tacbadge{ValColor}{Validation} Let’s take it slow. \tacbadge{AdvColor}{Advice}  What’s on your mind right now? \tacbadge{ExpColor}{Emotional Expression}\,\tacbadge{QueColor}{Questioning}}\par}

\method{\ourframework{}\minticon{}: Q\,+\,D$_{\text{KL}}$ (Ours)}{}
\modelsize{1.7B}{}
\scorecard{4.33}{4}{1.00}{4}{5}{2}{1}{1}{1}
{\raggedright\noindent {\small I can imagine how heavy this feels — like a quiet storm you didn’t expect to face. \tacbadge{ParColor}{Paraphrasing}\,\tacbadge{ValColor}{Validation} It’s completely okay to feel lost, especially when you’ve poured so much into this role. \tacbadge{ParColor}{Paraphrasing}\,\tacbadge{ValColor}{Validation} You’re not alone in this, and your worth isn’t tied to your job. \tacbadge{ValColor}{Validation} What does you feel when you think about your work now? \tacbadge{QueColor}{Questioning} Are there moments when you feel a spark of hope, even if it’s small? \tacbadge{ParColor}{Paraphrasing}\,\tacbadge{QueColor}{Questioning} You’re already doing something meaningful by nurturing those hobbies — maybe they’re quietly healing you. \tacbadge{ParColor}{Paraphrasing} It’s okay to pause and reorient. \tacbadge{ValColor}{Validation} You’ve got the strength to keep going, even when the path feels uncertain. \tacbadge{EwrColor}{Empowerment}\,\tacbadge{ParColor}{Paraphrasing} Would you like to share how you’re already showing up in your own way? Your resilience is a gift, and it’s worth celebrating. \tacbadge{EwrColor}{Empowerment}}\par}
\modelsize{4B}{}
\scorecard{4.83}{6}{1.00}{4}{5}{5}{1}{1}{1}
{\raggedright\noindent {\small It sounds like you’re feeling a mix of uncertainty and self-doubt about your future, especially after so many years in your role. \tacbadge{ParColor}{Paraphrasing} That’s completely normal — it’s not easy to shift from a familiar routine to something new, even when you’ve prepared for it. \tacbadge{ParColor}{Paraphrasing}\,\tacbadge{ValColor}{Validation} You’re not alone in feeling this way, and it’s okay to carry that weight. \tacbadge{ValColor}{Validation} What I hear is that you’re holding onto the idea that your hobbies are just "things you do for fun," but I wonder if there’s a way to see them as something more — like a foundation for new opportunities. \tacbadge{AdvColor}{Advice}\,\tacbadge{ExpColor}{Emotional Expression}\,\tacbadge{ParColor}{Paraphrasing} Would you feel comfortable sharing what gives you hope or what feels possible right now? \tacbadge{QueColor}{Questioning} I’m here to listen without judgment. You’ve already shown incredible strength by facing this challenge, and I believe you have the resilience to navigate this next chapter. \tacbadge{EwrColor}{Empowerment}\,\tacbadge{ParColor}{Paraphrasing} What would you say is one small step you could take toward feeling more in control? \tacbadge{QueColor}{Questioning}}\par}

\begin{findingsbox}
{\small The gold previous supporter turn used only \textit{questioning} (asking about hobbies). Vanilla immediately shifts to career advice (``update your resume, join LinkedIn groups''), treating the seeker's emotional distress as a problem to be solved with a to-do list. PsychoCounsel acknowledges feelings but still pivots to advice and information-giving. At 1.7B, \ourframework{}\minticon{} uses evocative language (``a quiet storm you did not expect to face'') and validates the seeker's worth as separate from their job (``your worth is not tied to your job''). At 4B, \ourframework{}\minticon{} connects the seeker's hobbies to a sense of purpose, though it does also suggest reframing hobbies as potential opportunities. Across both scales, \ourframework{}\minticon{} introduces three to five new tactics relative to the previous turn and shifts the conversation toward exploratory questioning rather than directive advice.}
\end{findingsbox}

\clearpage

\end{document}